\documentclass{article}

\usepackage{arxiv}

\usepackage{natbib}

\usepackage[utf8]{inputenc} 
\usepackage[T1]{fontenc}    
\usepackage{hyperref}       
\usepackage{url}            
\usepackage{booktabs}       
\usepackage{amsfonts}       
\usepackage{nicefrac}       
\usepackage{microtype}      
\usepackage{lipsum}
\usepackage{graphicx}

\usepackage{subfig}
\usepackage{wrapfig}
\usepackage[flushleft]{threeparttable}

\usepackage{algorithm}
\usepackage{algpseudocode}
\algdef{SE}[DOWHILE]{Do}{doWhile}{\algorithmicdo}[1]{\algorithmicwhile\ #1}%

\usepackage{hyperref}
\usepackage[symbol]{footmisc}
\usepackage{amsthm}
\usepackage{amsmath}   
{
      \theoremstyle{plain}
      \newtheorem{ass}{Assumption}
      \newtheorem{thm}{Theorem}
      \newtheorem{defn}{Definition}
      \newtheorem{lem}{Lemma}
      \newtheorem{cor}{Corollary}
}

\newcommand{\norm}[1]{\left\|#1\right\|}

\makeatletter
\def\blfootnote{\gdef\@thefnmark{}\@footnotetext}
\makeatother

\title{Hogwild! over Distributed Local Data Sets with Linearly Increasing Mini-Batch Sizes}

\author{Marten van Dijk$^{1,2*}$, Nhuong V. Nguyen$^{3\dagger*}$, Toan N. Nguyen$^{3\dagger}$, \\ \textbf{Lam M. Nguyen}$^{4}$\textbf{,} \textbf{Quoc Tran-Dinh}$^{5}$\textbf{,}  \textbf{Phuong Ha Nguyen}$^{6}$ \\
$^{1}$ CWI Amsterdam, The Netherlands\\
$^{2}$ 
Department of Electrical and Computer Engineering, University of Connecticut, CT, USA\\
$^{3}$ Department of Computer Science and Engineering, University of Connecticut, CT, USA \\
$^{4}$ IBM Research, Thomas J. Watson Research Center, Yorktown Heights, NY, USA\\
$^{5}$ Department of Statistics and Operations Research, \\
The University of North Carolina at Chapel Hill, Chapel Hill, NC, USA, \\
$^{6}$ eBay, CA, USA\\
\\
\texttt{marten.van.dijk@cwi.nl}, \texttt{nhuong.nguyen@uconn.edu}, \texttt{nntoan2211@gmail.com},  \\
\texttt{LamNguyen.MLTD@ibm.com}, \texttt{quoctd@email.unc.edu}, \texttt{phuongha.ntu@gmail.com}}

\begin{document}

\maketitle

\blfootnote{$^{*}$ these authors contributed equally.}
\blfootnote{$^{\dagger}$ supported by NSF grant CNS-1413996 “MACS: A Modular
Approach to Cloud Security.”}

\begin{abstract}
Hogwild! implements asynchronous Stochastic Gradient Descent (SGD) where multiple threads in parallel access a common repository containing training data, perform SGD iterations, and update  shared state that represents a jointly learned (global) model. We consider big data analysis where training data is distributed among local data sets in a heterogeneous way -- and we wish to move SGD computations to local compute nodes where local data resides. The results of these local SGD computations are aggregated by a central ``aggregator'' which mimics Hogwild!. We show how local compute nodes can start choosing small mini-batch sizes which increase to larger ones in order to reduce communication cost (round interaction with the aggregator). 
We improve state-of-the-art literature and show  $O(\sqrt{K}$) communication rounds for heterogeneous data for strongly convex problems, where $K$ is the total number of gradient computations across all local compute nodes.
For our scheme, we prove a \textit{tight} and novel non-trivial convergence analysis for strongly convex problems for {\em heterogeneous} data which does not use the bounded gradient assumption as seen in many existing publications. The tightness is a consequence of  our proofs for  lower and upper bounds of the convergence rate, which show a constant factor difference. We show experimental results for plain convex and non-convex problems for biased (i.e., heterogeneous)  and unbiased local data sets.
\end{abstract}

\section{Introduction}

The optimization problem for training many Machine Learning (ML) models using a training set $\{\xi_i\}_{i=1}^M$ of $M$ samples can be formulated as a finite-sum minimization problem as follows
$$ 
\min_{w \in \mathbb{R}^d} \left\{ F(w) = \frac{1}{M}
\sum_{i=1}^M f(w; \xi_i) \right\}.
$$
The objective is to minimize a loss function with respect to model parameters $w$. This problem is known as empirical risk minimization and it covers a wide range of convex and non-convex problems from the ML domain, including, but not limited to, logistic regression, multi-kernel learning, conditional random fields and neural networks.
In this paper, we are interested in solving the following more general stochastic optimization problem with respect to some distribution $\mathcal{D}$:
\begin{align}
\min_{w \in \mathbb{R}^d} \left\{ F(w) = \mathbb{E}_{\xi \sim \mathcal{D}} [ f(w;\xi) ] \right\},  \label{eqObj}  
\end{align}
where $F$ has a Lipschitz continuous gradient and $f$ is bounded from below for every $\xi$.

Big data analysis in the form of ML over a large training set distributed over local databases requires computation to be moved to compute nodes where local training data resides. Local SGD computations are communicated to a central ``aggregator'' who maintains a global model. Local computations are executed in parallel and resulting SGD updates arrive out-of-order at the aggregator. For this purpose we need robust (in terms of convergence) asynchronous SGD.

Our approach is based on the Hogwild!~~\citep{Hogwild}
recursion 
\begin{equation}
 w_{t+1} = w_t - \eta_t  \nabla f(\hat{w}_t;\xi_t),\label{eqwM2a}
 \end{equation}
 where $\hat{w}_t$ represents the vector used in computing the gradient $\nabla f(\hat{w}_t;\xi_t)$ and whose vector entries have been read (one by one)  from  an aggregate of a mix of  previous updates that led to $w_{j}$, $j\leq t$.
 In a single-thread setting where updates are done in a fully consistent way, i.e. $\hat{w}_t=w_t$, yields SGD with diminishing step sizes $\{\eta_t\}$. 

Recursion (\ref{eqwM2a}) models asynchronous SGD.
We define the amount of asynchronous behavior by function $\tau(t)$:

\begin{defn}
We say that the sequence $\{w_t\}$  is {\em consistent with a delay function $\tau$}  
if, for all $t$, vector $\hat{w}_t$ includes the aggregate 
of the updates up to and including those made during the $(t-\tau(t))$-th iteration\footnote{ (\ref{eqwM2a}) defines the $(t+1)$-th iteration, where $\eta_t\nabla f(\hat{w}_t;\xi_t)$ represents the $(t+1)$-th update.}, i.e., $\hat{w}_t = w_0 - \sum_{j\in {\cal U}} \eta_j \nabla f(\hat{w}_j;\xi_j)$ for some ${\cal U}$ with $\{0,1,\ldots, t-\tau(t)-1\}\subseteq {\cal U}$.
\end{defn}

Our main insight is that the asynchronous SGD framework based on Hogwild! can resist much larger delays than the natural delays caused by the network communication infrastructure, in fact, it turns out that $\tau(t)$ can scale as much as $\approx \sqrt{t/\ln t}$ for strongly convex problems \citep{nguyen2018sgd,nguyen2018new}. This means that recurrence (\ref{eqwM2a}) can be used/exploited in an asynchronous SGD implementation over distributed local data sets where much more {\em asynchronous behavior is introduced  by design}.


In our setting where SGD recursions are executed locally at compute nodes where biased (i.e., heterogeneous) local training data resides, compute nodes execute SGD recursions in `rounds'. From
the perspective of a local compute node, a round consists of the sequence of SGD recursions 
between two consecutive `update' communications to the central aggregator. Within a round, $w_t$ is iteratively updated by subtracting $\eta_t \nabla f(\hat{w}_t;\xi_t)$. The sum of updates $\sum_{t \in {\sc round}} \eta_t \nabla f(\hat{w}_t;\xi_t)$ is communicated to the aggregator at the end of the round after which a new round starts leading to a next sum of updates  to be communicated at the end of the next round. Locally, compute nodes receive (out of sync)  broadcast messages with the current global model from the central aggregator (according to some strategy, e.g., on average one or two such messages per round). This is used to replace their local model $w_t$ computed so far. Details are given in Section \ref{sec:alg}. 


Rather than having each round  execute the same/constant number of SGD recursions, 
this paper builds on our main observation that we are allowed to introduce asynchronous behavior by design while still maintaining convergence (as we will see at a rate  that matches a \textit{lower bound} up to a constant). We propose to {\em increase the number of SGD iterations performed locally at each compute node  from round to round}.
This {\em reduces the amount of network communication} compared to the straightforward usage of recurrence (\ref{eqwM2a}) where each compute node performs a fixed number of SGD iterations within each round.

In the distributed data setting, each local compute node may only execute for a fraction of an epoch (where the number of iterations, i.e., gradient computations, in an epoch is equal to the size of the global big training data set defined as the collection of all local training data sets together). In order to disperse to all local compute nodes
information from local updates by means of updating the global model and receiving feedback from the global model (maintained at the central aggregator), there needs to be a sufficient number of round interactions. For the best convergence, we need to have more round interactions at the very start where local updates contain the most (directional) information about (where to find) the global minimum to which we wish to converge. This corresponds to small 'sample' sizes (measured in the number of local SGD updates within a round) in the beginning. And in order to gain as much useful information about where to find the global minimum, initial local updates should use larger step sizes (learning rate).

From our theory 
we see that in order to bootstrap convergence it is indeed the best  to start with larger step sizes and start with  rounds of small sample size 
after which these rounds should start  increasing in  sample size and should start using smaller and smaller step sizes 
for best performance (in terms of minimizing communication while still achieving high test accuracy).
Experiments 
confirm our expectations. 


\noindent
{\bf Contributions.} For the distributed data setting where SGD recursions are executed locally at compute nodes where local training data resides and which update a global model maintained at a centralized aggregator,  
we introduce a new SGD algorithm based on Hogwild! \citep{Hogwild} which does not use fixed-sized min-batch SGD at the local compute nodes but uses increasing mini-batch (sample) sizes from round interaction to round interaction:

 {\bf [I]}    The compute nodes and server can work together to create a global model 
     in asynchronous fashion, where  we assume that 
messages/packets never drop; they will be re-sent but can arrive out of order. In   Theorem \ref{thmalg} we characterize distribution ${\cal D}$ in the stochastic optimization problem (\ref{eqObj}) to which the global model relates.  

{\bf [II]} Given a specific (strongly convex, plain convex or non-convex) stochastic optimization problem, we may assume (believe in) a delay function $\tau$ which characterizes the maximum asynchronous behavior which our algorithm can resist for the specific problem.
Given the delay function $\tau$, we provide a general recipe for constructing increasing sample size sequences and diminishing round step size sequences so that our algorithm maintains $\tau$ as an invariant.
 For strongly convex problems, \citep{nguyen2018sgd,nguyen2018new} prove that
   $\tau(t)$ can be as large as $\approx \sqrt{t / \ln t}$ for which our recipe shows a diminishing round step size sequence of $O(\frac{\ln i}{i^2})$, where $i$ indicates the round number, that  allows a sample size sequence of $\Theta(\frac{i}{\ln i})$;  the sample size sequence can almost {\em linearly} increases from round to round.
   


    {\bf [III]} 
  For strongly convex problems with `linearly' increasing sample size sequences $\Theta(\frac{i}{\ln i})$
   we  prove in Theorem \ref{thmsc2} an upper bound of $O(1/t)$ on the expected convergence rate $\mathbb{E}[\|w_{t} - w_* \|^2]$, where $w_*$ represents the global minimum in (\ref{eqObj}) and $t$ is the SGD iteration number (each local node computes a subset of the $w_t$). 
In fact, the concrete  expression of the upper bound  attains for increasing $t$ the best possible convergence rate (among stochastic first order algorithms)  within a constant factor $\leq 8\cdot 36^2$, see Corollary \ref{lem:lower} which directly applies the lower bound from \citep{nguyen2019tight}.

    {\bf [IV]} 
Let $K$ be the total number of gradient computations (summed over all local nodes)  needed for the desired test accuracy, and let $T$ be the number of communication rounds in our algorithm. Then, $T$ scales less than linear with $K$ due to the increasing sample size sequence (if a constant sample size sequence is used, i.e., fixed-sized mini-batch SGD at each of the local compute nodes, then $T=O(K)$).
For strongly convex problems with diminishing step sizes we show $T= O(\sqrt{K})$ for heterogeneous local data while having $O(1/K)$ convergence rate. This implies that using diminishing step sizes give a much better performance compared to constant step sizes in terms of communication.


{\bf [V]} Experiments for linearly increasing sample size sequences for strongly convex problems as well as plain convex and non-convex problems confirm the robustness of our algorithm in terms of good test accuracies (our theoretical understanding from strongly convex problems seems to generalize to plain and non-convex problems). We use biased local training data sets (meaning different compute nodes see differently biased subsets of training data) and compare to unbiased local training data sets.

\section{Related Work}

\noindent
{\bf Unbiased Local Data (iid).}
For strongly convex problems with unbiased local data sets, ~\citep{stich2018local} showed $O(1/K)$ convergence rate for $O(\sqrt{K})$ communication rounds. For the iid case this was improved by~\citep{spiridonoff2020local} to just $1$ communication round, where each client performs local SGD separately after which in ``one shot'' all local models are aggregated (averaged) -- this corresponds to $O(n)$ total communication for $n$ clients. This result was generalized by~\citep{khaled2020tighter}. Based on the strong Polyak-Lojasiewicz (PL) assumption (which is a generalization of strong convexity but covers certain nonconvex models), \citep{yu2019computation} proved for the iid case  a convergence rate  of $O(1/K)$ with $\log(K/n)$ communication rounds with an exponentially increasing sample size sequence.
For non-convex problems,
\citep{yu2019computation}
proved for the iid case the standard convergence rate $O(1/\sqrt{K})$ (as defined for non-convex problems) with $O(\sqrt{K}\log (K/n^2))$ communication rounds.

\noindent
{\bf Biased Local Data (heterogeneous).} Our focus is on heterogeneous data between different clients (this needs more communication rounds in order to achieve convergence, one-shot averaging is not enough): \citep{khaled2020tighter} were the first to analyze the convergence rate for plain convex problems in this scenario. They use the bounded variance assumption in their analysis with constant step-size and with sample size sequences where sample sizes are bound by an a-priori set parameter $H$. They prove that $O(1/\sqrt{K})$ convergence rate (optimal for plain convex) is achieved for $O(K^{3/4})$ communication rounds (see their Corollary $5$ and notice that their algorithm uses $(K/n)/H$ communication rounds). 
For strongly convex problems in the {\em heterogeneous} case (without assuming bounded variance), we show that {\em convergence rate $O(1/K)$ (optimal for strongly convex) is achieved for $O(\sqrt{K})$ communication rounds}. 



\noindent 
{\bf Asynchronous Training.}
Asynchronous training~\citep{zinkevich2009slow,lian2015asynchronous,zheng2017asynchronous,meng2017asynchronous,stich2018local,shi2019distributed} is widely used in traditional distributed SGD. Hogwild!, one of the most famous asynchronous SGD algorithms, was introduced in ~\citep{Hogwild} and various variants with a fixed and diminishing step size sequences were introduced in \citep{ManiaPanPapailiopoulosEtAl2015,DeSaZhangOlukotunEtAl2015,Leblond2018,nguyen2018sgd}. 
%
Typically, asynchronous training converges faster than synchronous training in real time due to parallelism. This is because in a synchronized solution compute nodes have to wait for the slower ones to communicate their updates after which a new global model can be downloaded by everyone. This causes high idle waiting times at compute nodes.
Asynchronous training allows compute nodes to continue executing SGD recursions based on stale global models. For non-convex problems, synchronous training \citep{Ghadimi2013Minibatch} and  asynchronous training with bounded staleness \citep{lian2015asynchronous}, or in our terminology bounded delay, achieves the same convergence rate of $O(1/\sqrt{K})$, where  $K$ is the total number of gradient computations. 


The methods cited above generally use mini-batch SGD (possibly with diminishing step sizes from round to round) at each of the distributed computing threads, hence, 
parallelism will then lead to asynchronous behavior dictated by a delay $\tau$ which can be assumed to be bounded. Assuming bounded delays, the convergence rate is mathematically analysed in the papers cited above with the exception of \citep{nguyen2018sgd,nguyen2018new} which also analyses the convergence rate for unbounded delays (i.e., increasing delay functions $\tau$).

As explained in this introduction, \textit{this paper exploits the advantage of being able to resist much more asynchronous behavior than bounded delay}. We show how one can use diminishing step sizes alongside increasing sample sizes (mini-batch sizes) from round to round. This provides a technique complimentary to~\citep{zinkevich2009slow,lian2015asynchronous,lian2017asynchronous,zheng2017asynchronous,meng2017asynchronous,stich2018local,shi2019distributed} with which current asynchronous training methods can be enhanced at the benefit of reduced communication -- which is important when training over distributed local data sets in big data analysis. For example,
rather than sending gradients to the server after each local update, which is not practical for edge devices (such as mobile or IoT devices) due to  unreliable and slow communication,
\citep{shi2019distributed} introduces the idea of using a tree like communication structure which aggregates local updates in pairs from leafs to root -- this technique for meeting throughput requirements at the centralized server can be added to our technique of increasing sample sizes from round to round. As another example, \citep{lian2017asynchronous} introduces asynchronous decentralized SGD where local compute nodes do not communicate through a centralized aggregator but instead perform a consensus protocol -- our technique of increasing sample sizes is complementary and can possibly be beneficial to use in this decentralized network setting. Similarly, our technique may apply to \citep{jie2020asynchronous}, where the authors studied a new asynchronous decentralized SGD with the goal of offering privacy guarantees. 
We stress that our setting is asynchronous centralized SGD and  is completely different from asynchronous decentralized SGD algorithms as in~\citep{shi2019distributed,lian2017asynchronous,jie2020asynchronous}.

\noindent 
{\bf Federated Learning and Local SGD.}
Federated Learning (FL) \citep{jianmin,mcmahan} is a distributed machine learning approach which enables training on a large corpus of decentralized data located on devices like mobile phones or IoT devices. Federated learning brings the concept of ``bringing the code to the data, instead of the data to the code" \citep{bonawitz2019towards}. Google \citep{konevcny2016federated} demonstrated FL for the  first time at a large scale  when they conducted experiments of training a global model across all mobile devices via the Google Keyboard Android application \citep{GoogleAIBlog}.

Original FL requires synchrony between the server and clients (compute nodes).  It requires that each client sends a full model back to the server in each round and each client needs to wait for the next computation round. For large and complicated models, this becomes a main bottleneck  due to the asymmetric property of internet connection and the different computation power of devices \citep{yang,konevcny2016federated,luping,kevin}.

In \citep{cong2019,yang2019} asynchronous training combined  with federated optimization is proposed. Specifically, the server and workers (compute nodes) conduct updates asynchronously: the server immediately updates the global model whenever it receives a local model from clients. Therefore, the communication between the server and workers is non-blocking and more effective. 
We notice that~\citep{cong2019} provides a convergence analysis, while~\citep{yang2019} does not.

In \citep{tianli2019hetero}, the authors introduce FedProx which is a modification of FedAvg (i.e., original FL algorithm of \citep{mcmahan}). In FedProx, the clients solve a proximal minimization problem rather than traditional minimization as in FedAvg. For theory, the authors use $B$-local dissimilarity and bounded dissimilarity assumptions for the global objective function. This implies that there is a bounded gradient assumption applied to the global objective function. Moreover, their proof requires the global objective function to  be strongly convex.

One major shortcoming in the terms of convergence analysis of asynchronous SGD in many existing publications is that the bounded gradients and strongly convex assumptions are used together, e.g.   ~\citep{lian2015asynchronous,lian2017asynchronous,ManiaPanPapailiopoulosEtAl2015,DeSaZhangOlukotunEtAl2015,jie2020asynchronous,cong2019,mcmahan,tianli2019hetero}. However, the bounded gradient assumption is in conflict with assuming strong convexity as explained in \citep{nguyen2018sgd,nguyen2018new}. It implies that the convergence analysis should not use these two assumptions together to make the analysis complete.

Our method of increasing sample sizes together with its convergence analysis for strongly objective functions (which does not use the bounded gradient assumption) complements the related work in FL and local SGD. This paper analyses and demonstrates the promise of this new technique but does not claim a full end-to-end implementation of FL or distributed SGD with asynchronous learning. 


\section{Hogwild! \& Increasing Sample Sizes}
\label{sec:alg}

The next subsections explain our proposed algorithm together with how to set parameters in terms of a concrete round to round diminishing step size sequence and increasing sample size sequence.

\subsection{Asynchronous SGD over Local Data Sets}

\begin{algorithm}[!ht]
\caption{ComputeNode$_c$ -- Local Model}
\label{alg:DP}

\begin{algorithmic}[1]
\Procedure{Setup}{$n$}: Initialize  increasing sample size sequences $\{s_i\}_{i\geq 0}$ and $\{s_{i,c}\approx p_cs_i\}_{i\geq 0}$ for each compute node $c\in \{1,\ldots, n\}$, where $p_c$ scales the importance of compute node $c$. Initialize diminishing round step sizes $\{\bar{\eta}_i\}_{i\geq 0}$, a permissible delay function $\tau(\cdot)$ with $t-\tau(t)$ increasing in $t$, and a default global model for each compute node to start with.
\EndProcedure   
\State

\Procedure{ISRReceive}{$\hat{v},k$}: This Interrupt Service Routine is called whenever a broadcast message with a new global model $\hat{v}$ is received from the server. Once received, the compute node's local model $\hat{w}$ is replaced with $\hat{v}$
from which the latest accumulated update $\bar{\eta}_i \cdot U$ of the compute node in the ongoing round (as maintained in line 17 below)   is subtracted. 
(We notice that 
the $k$th broadcast message containing a global model $\hat{v}$ from the server is transmitted by the server as soon as 
the updates up to and including rounds $0, \ldots, k-1$  from {\em all} compute nodes have been received; thus
$\hat{v}$ 
includes all the updates up to and including round $k-1$.)
\EndProcedure   

\State

\Procedure{MainComputeNode}{$\mathcal{D}_c$} 
\State $i=0$, $\hat{w}=\hat{w}_{c,0,0}$ 
 
\While{\textbf{True}}
    \State $h=0$, $U=0$ 
    \While{$h< s_{i,c}$}
    
              \State $t_{glob}= s_0+\ldots+s_{i} - (s_{i,c}-h)-1$ 
       \State $t_{delay} = s_{k}+\ldots+s_{i} - (s_{i,c}-h) $
       \State {\bf while} 
       $\tau(t_{glob}) < t_{delay}$ 
       {\bf do} nothing 

        \State Sample uniformly at random $\xi$ from $\mathcal{D}_c$ 
        
        \State $g = \nabla f(\hat{w}, \xi)$ 
        
        
        \State ${U} = {U} + g$  
        
        \State Update model $w = \hat{w} - \bar{\eta}_{i} \cdot g$ 
        \State \Comment{$w$ represents $ w_{c,i,h+1}$} 
         \State Update model $\hat{w} = w$ 


        \State $h$++
    \EndWhile
    
    \State Send $(i,c, U)$ to the Server. 
    \State $i$++
\EndWhile
   
\EndProcedure

\end{algorithmic}
\end{algorithm}

Compute node $c\in \{1,\ldots,n\}$ updates its local model $\hat{w}$ according to
Algorithm \ref{alg:DP}. Lines 15, 16, 18, and 20 represent an SGD recursion where $\xi$ is sampled from distribution ${\cal D}_c$, which represents $c$'s local data set. Variable $U$, see line 17, keeps track of the sum of the gradients that correspond to $s_{i,c}$ samples during $c$'s $i$-th local round.  This information is send to the server in line 23, who will multiply $U$ by the round step size $\bar{\eta}_i$ and subtract the result from the global model $\hat{v}$. In this way each compute node contributes updates $U$ which are aggregated at the server. As soon as the server has aggregated each compute node's updates for their first $k$ local rounds, the server broadcasts global model $\hat{v}$. As soon as $c$ receives $\hat{v}$ it replaces its local model $\hat{w}$ with $\hat{v}-\bar{\eta}_i\cdot U$ (this allows the last computed gradients in $U$ that have not yet been aggregated into the global model at the server not to go to waste). 

Line 14 shows that a compute node $c$ will wait until $t_{delay}$ becomes smaller than $\tau(t_{glob})$. This will happen as a result of \Call{ISRReceive}{$\hat{v},k$} receiving broadcast message $\hat{v}$ together with a larger $k$ after which the ISR computes a new (smaller) $t_{delay}$. Line 14 guarantees the invariant $t_{delay}\leq \tau(t_{glob})$, where $\tau$ is initialized to some "permissible" delay function which characterizes the amount of asynchronous behavior we assume the overall algorithm can tolerate (in that the algorithm has fast convergence leading to good test accuracy). 
  
Supplemental Material \ref{app-alg} has all the detailed pseudo code with annotated invariants.

We want to label the SGD recursions computed in each of the \Call{MainComputeNode}{${\cal D}_c$} applications for $c\in \{1,\ldots, n\}$ to an iteration count $t$ that corresponds to recursion (\ref{eqwM2a}): We want to put all SGD recursions computed by each compute node in sequential order such that it is as if we used (\ref{eqwM2a}) on a single machine. This will allow us to analyse whether our algorithm leads to a sequence $\{w_t\}$ which is consistent with the initialized permissible delay function $\tau$. In  order to find an ordering based on $t$ we define in Supplemental Material \ref{appmain} a mapping $\rho$ from the annotated labels $(c,i,h)$ in \Call{MainComputeNode}{} to $t$ and use this to prove the following theorem:

\begin{thm} \label{thmalg}
Our setup, compute node, and server algorithms produce a sequence $\{w_t\}$ according to recursion (\ref{eqwM2a}) where $\{\xi_t\}$ are selected from distribution ${\cal D}=\sum_{c=1}^n p_c {\cal D}_c$. Sequence $\{w_t\}$ is consistent with delay function $\tau$ as defined in \Call{Setup}{}.
\end{thm}

The theorem tells us that the algorithms implement  recursion (\ref{eqwM2a}) for distribution ${\cal D}=\sum_{c=1}^n p_c {\cal D}_c$, i.e., a convex combination of each of the (possibly biased) local distributions (data sets). Scaling factors $p_c$ represent a distribution (i.e., they sum to 1) and are used to compute local sample sizes $s_{i,c}\approx p_c s_i$, where $s_i= \sum_{c=1}^n s_{i,c}$ indicates the total number of samples in rounds $i$ across all compute nodes.

We remark that our asynchronous distributed SGD is compatible with the more general recursion mentioned in \citep{nguyen2018sgd,nguyen2018new} and explained in Supplemental Material \ref{app-rec}. In this recursion each client can apply a "mask" which indicates the entries of the local model that will be considered. This allows each client to only transmit the local model entries corresponding to the mask.

\subsection{Delay $\tau$}

We assume that 
messages/packets never drop. They will be resent 
but can arrive out-of-order. We are robust against this kind of asynchronous behavior:
The amount of asynchronous behavior is limited by $\tau(\cdot)$; when the delay is getting too large, then the client (local compute node) enters a wait loop which terminates only when \Call{ISRReceive}{} receives a more recent global model $\hat{v}$ with higher $k$ (making $t_{delay}$ smaller). Since $\tau(t)$ increases in $t$ and is much larger than the delays caused by network latency and retransmission of dropped packets, asynchronous behavior due to such effects will not cause clients to get stuck in a waiting loop. We assume different clients have approximately the same speed of computation which implies that this will not cause fast clients having to wait for long bursts of time.\footnote{When entering a waiting loop, the client's operating system should context switch out and resume other computation. 
If a client $c$ is an outlier with slow computation speed, then we can adjust $p_c$ to be smaller in order to have its mini-batch/sample size $s_{i,c}$ be proportionally smaller; this will change distribution ${\cal D}$ and therefore change the objective function of (\ref{eqObj}).
} 

We exploit the algorithm's resistance against delays $\tau(t)$ by using increasing sample size sequences $\{s_{i,c}\}$. Since the server only broadcasts when all clients have communicated their updates for a "round" $k$, increasing sample sizes implies that $t_{delay}$ can get closer to $\tau(t_{glob})$. So, sample size sequences should not increase too much: We require the property that there exists a threshold $d$ such that for all $i\geq d$,
\vspace{-.2cm}
\begin{equation} \tau \left(\sum_{j=0}^i s_j \right )\geq 1+\sum_{j=i-d}^i s_j. \label{eqtausample}
\end{equation}
In Supplemental Material \ref{appinv} we show that this allows us to replace  condition $\tau(t_{glob})<t_{delay}$ 
of the waiting loop by 
$i>k+d$ when $i\geq d$
while still guaranteeing $t_{delay}\leq \tau(t_{glob})$ as an invariant. In practice, since sample sizes increase, we only need to require (\ref{eqtausample}) for $d=1$ (which means we allow a local lag of one communication round) in order to resist asynchronous behavior due to network latency.

\subsection{Recipe Sample Size Sequence}

Given a fixed budget/number of gradient computations $K$ which the compute nodes together need to perform, an increasing sample size sequence $\{s_i\}$ reduces the number $T$ of communication rounds/interactions (defined by update messages coming from the compute nodes with broadcast messages from the server). Convergence to an accurate solution must be guaranteed, that is, $\{s_i\}$ has to satisfy (\ref{eqtausample}) if we assume $\tau$ is indeed a permissible delay function.

Supplemental Material \ref{appinc} proves how a general formula for function $\tau$ translates into an increasing sample size sequence $\{s_i\}$ that satisfies (\ref{eqtausample}). 

\begin{lem} \label{lemsample-simple}
Let $g>1$.
Suppose that 
$\tau(x) = M_1 +   (x+M_0)^{1/g}$
for some  $M_1\geq d+2$ and
$M_0 \geq ((m+1)(g-1)/g)^{g/(g-1)}$, where $m\geq 0$ is an integer. Then
$$s_i = \left\lceil \frac{1}{d+1}\left(\frac{m+i+1}{d+1}\frac{g-1}{g} \right)^{1/(g-1)} \right\rceil$$ satisfies property (\ref{eqtausample}).
\end{lem}

The above lemma is a direct consequence of Supplemental Material \ref{appinc} which has a more general proof that also allows functions such as $\tau(x)= M_1 +   ((x+M_0)/\ln (x+M_0))^{1/g}$ (needed for analysing the convergence of strongly convex problems with $g=2$).

\subsection{Recipe Round Step Size Sequence}

As soon as we have selected an increasing sample size sequence based on $\tau$, Supplemental Material \ref{appstep} shows how we can translate the diminishing step size sequence $\{\eta_t\}$ of recurrence (\ref{eqwM2a}) to a diminishing round step size sequence $\{\bar{\eta}_i\}$ that only diminishes with every mini-batch $s_i$ from round to round. The lemma below is a direct consequence of Supplemental Material \ref{appstep} which has a slightly more general statement.

\begin{lem} \label{lemstep-simple}
Let $0\leq q\leq 1$ and $\{E_t\}$ a constant or increasing sequence with $E_t\geq 1$. For $q$ and $\{E_t\}$ consider the set ${\cal Z}$ of diminishing step size sequences $\{\eta_t\}$
in recurrence (\ref{eqwM2a}) with  
$\eta_t = \alpha_t / (\mu (t+E_t)^q)$ where $\{\alpha_t\}$ is some  sequence of values with $\alpha_0\leq \alpha_t\leq 3\cdot \alpha_0$.

We assume sample size sequence $\{s_i\}$ of Lemma \ref{lemsample-simple} for $g\geq 2$.
For $i\geq 0$, we define
$\bar{E}_i = E_{\sum_{j=0}^i s_j}$. We define $\bar{E}_{-1}=E_0$.
If $\bar{E}_{i} \leq 2 \bar{E}_{i-1}$ for $i\geq 0$ and if
$s_0-1\leq E_0$, then there exists a 
 diminishing step size sequence $\{\eta_t\}$ in set ${\cal Z}$ such that
$$\eta_t = \frac{\alpha_t}{\mu(t+E_t)^q} = \frac{\alpha_0}{\mu ((\sum_{j=0}^{i-1} s_j) +\bar{E}_{i-1})^q} 
\stackrel{\mbox{{\footnotesize {\sc def}}}}{=} \bar{\eta}_i
$$ 
for $t\in \{(\sum_{j=0}^{i-1} s_j), \ldots, (\sum_{j=0}^{i-1} s_j) +s_i -1 \}$.
\end{lem}

Notice $s_i = \Theta(i^{1/(g-1)})$ and 
$\bar{\eta_i}=O(i^{-q\cdot (1+1/(g-1))})$.

In Supplemental Material \ref{app-nonconvex} we discuss plain and non-convex problems. We argue  in both cases to choose  a  diminishing step size sequence of $O(t^{-1/2})$, i.e., $q=1/2$, and to experiment with different increasing sample size sequences $\Theta(i^{1/(g-1)})$, for $g\geq 2$, to determine into what extent the presented asynchronous SGD  is robust against delays. Substituting $p=1/(g-1)\in (0,1]$ gives sample size sequence $\Theta(i^{p})$ and round step size sequence $O(i^{-(1+p)/2})$. It turns out that $p=1$ gives a good performance and this confirms our intuition that the results from our theory on strongly convex functions in the next section generalizes in that also plain and non-convex problems have fast convergence for linearly increasing sample size sequences.

\section{Convergence Rate for Strongly Convex problems}

In this section we provide a round step size sequence and a sample size sequence for strongly convex problems. We show tight upper and lower bounds.
For strongly convex problems, we assume the following:

\begin{ass}[$L$-smooth]
\label{ass_smooth}
$f(w;\xi)$ is $L$-smooth for every realization of $\xi$, i.e., there exists a constant $L > 0$ such that, $\forall w,w' \in \mathbb{R}^d$, 
\begin{align*}
\| \nabla f(w;\xi) - \nabla f(w';\xi) \| \leq L \| w - w' \|. \label{eq:Lsmooth_basic}
\end{align*} 
\end{ass}

\begin{ass}\label{ass_convex}
$f(w;\xi)$ is convex for every realization of $\xi$, i.e., $\forall w,w' \in \mathbb{R}^d$, 
\begin{gather*}
f(w;\xi)  - f(w';\xi) \geq \langle \nabla f(w';\xi),(w - w') \rangle.
\end{gather*}
\end{ass}

\begin{ass}[$\mu$-strongly convex]
\label{ass_stronglyconvex}
The objective function $F: \mathbb{R}^d \to \mathbb{R}$ is a $\mu$-strongly convex, i.e., there exists a constant $\mu > 0$ such that $\forall w,w' \in \mathbb{R}^d$, 
\begin{gather*}
F(w) - F(w') \geq \langle \nabla F(w'), (w - w') \rangle + \frac{\mu}{2}\|w - w'\|^2. \label{eq:stronglyconvex_00}
\end{gather*}
\end{ass}

Being strongly convex implies that $F$ has a global minimum $w_{*}$. For $w_{*}$ we assume:

\begin{ass}[Finite $\sigma$]
\label{ass_finitesigma}
Let $N = 2 \mathbb{E}[ \|\nabla f(w_{*}; \xi)\|^2 ]$ where $w_{*} = \arg \min_w F(w)$. We require $N < \infty$. 
\end{ass}

In this section we 
let $f$ be $L$-smooth, convex, and let the objective function $F(w)=\mathbb{E}_{\xi\sim {\cal D}}[f(w;\xi)]$ be $\mu$-strongly convex with finite $N = 2 \mathbb{E}[ \|\nabla f(w_{*}; \xi)\|^2 ]$ where $w_{*} = \arg \min_w F(w)$.
Notice that we do not assume the bounded gradient assumption which assumes $\mathbb{E}[ \|\nabla f(w; \xi)\|^2 ]$ is bounded for all $w\in \mathbb{R}^d $ (not only $w=w_{*}$ as in Assumption \ref{ass_finitesigma}) and is in conflict with assuming strong convexity as explained in \citep{nguyen2018sgd,nguyen2018new}.

\subsection{Sample Size and Round Step Size Sequences}

After sequentially ordering the SGD recursions from all compute nodes we end up with a Hogwild! execution as defined by recursion (\ref{eqwM2a}) and we may apply the results from \citep{nguyen2018sgd,nguyen2018new} that state that $\{w_t\}$ is consistent with any delay function $\tau(t)\leq \sqrt{(t/\ln t) \cdot (1- 1/\ln t)}$. By suitably choosing such a function $\tau$ (see Supplemental Material \ref{app-sc-dim} for details), application of the more general Lemma \ref{lemsample-simple} from Supplemental Material \ref{appinc} gives sample size sequence
$$s_i = \left \lceil \frac{m+i+1}{16(d+1)^2} \frac{1}{ \ln(\frac{m+i+1}{2(d+1)})} \right \rceil =
\Theta \left(\frac{i}{\ln i} \right ).$$

For a fixed number of gradient computations $K$, the number $T$ of communication rounds satisfies $K=\sum_{j=0}^T s_j$. When forgetting the $\ln i$ component, this makes $T$ proportional to $\sqrt{K}$ -- rather than proportional to $K$ for a constant sample size sequence. This  reduction in communication rounds and overall network communication  is possible because we use a diminishing step size sequence (which allows us to still prove tight upper and lower bounds  on the convergence rate).

For our choice of $\tau$, we are restricted in the more general Lemma \ref{lemstep-simple} from Supplemental Material \ref{appstep} to a family ${\cal Z}$ of step size functions with $a_0=12$ and $E_t=2\tau(t)$. For strongly convex problems we may choose $q=1$ which gives a step size sequence $\eta_t = \alpha_t / (\mu (t+E_t)^q)$ for which the convergence rate $\mathbb{E}[\|w_{t} - w_* \|^2]=O(1/t)$. This results in a $O \left (\frac{\ln i}{i^2} \right)$ round step size sequence $$
\bar{\eta}_i = \frac{12}{\mu} \cdot \frac{1}
{\sum_{j=0}^{i-1} s_j + 
2M_1
+
\sqrt{\frac{(m+1)^2/4+\sum_{j=0}^{i-1} s_j}{\ln((m+1)^2/4+ \sum_{j=0}^{i-1} s_j)}}},
%
$$
where
$$ M_1=\max \left \{d+2, 
72\cdot \frac{L}{\mu}, 
\frac{1}{2}\left\lceil \frac{m+1}{16(d+1)^2} \frac{1}{ \ln(\frac{m+1}{2(d+1)})} \right\rceil
\right \}$$

\subsection{Upper Bound Convergence Rate}


Based on the sequences $\{s_i\}$ and $\{\bar{\eta}_i\}$ we prove in Supplemental Material \ref{app-sc-dim} the following upper bound on the convergence rate for strongly convex problems:\footnote{We remark that this theorem also holds for our algorithm where the compute nodes compute mini-batch SGD for the sample set and it can be adapted for the more general recursion with masks as explained in Supplemental Material \ref{app-rec} (with ``$D>1$'').}

\begin{thm} \label{thmsc2}
For sample size sequence $\{s_i\}$ and round step size sequence $\{\bar{\eta}_i\}$ we have expected convergence rate
\begin{equation}
\mathbb{E}[\|w_{t} - w_* \|^2]\leq \frac{4\cdot 36^2 \cdot N}{\mu^2} \frac{1}{t}
+ O \left(\frac{1}{t \ln t} \right ).
\label{eqUCR}
\end{equation}
where $t$ represents the total number of gradient evaluations over all compute nodes performed so far.\footnote{Mapping $\rho$ maps annotated labels to $t$: $w_t=w_{\rho(c,i,h)}$.}
\end{thm}

Notice that $t$ is equal to  $n$ times  the average number of grad evaluations $\bar{t}$ per compute node, hence, convergence rate $O(1/t)=O(1/(n\bar{t}))$ showing the expected $1/n$ dependence. We also remind the reader that $s_i=\sum_c s_{i,c}$ where $s_{i,c}\approx p_c s_i$.

We do not know whether the theory for delay functions $\tau$ for strongly convex problems gives a tight bound in terms of the maximum delay\footnote{Permissible delay functions can possibly be larger than $\sqrt{(t/\ln t)\cdot (1- 1/\ln t)}$.} for which we can prove a tight upper bound on the convergence rate. For this reason we also experiment with the larger linear $s_i =\Theta(i)$ in the strongly convex case. (For $s_i=\Theta(i)$ we have $\bar{\eta}_i= O(i^{-2})$.)

As one benchmark we compare to using a constant step size $\eta=\bar{\eta}_i$. 
Supplemental Material \ref{appconst} analyses this case and shows how to choose the constant sample size $s=s_i$ (as large as $\frac{a}{ L\mu (d+1)}$ for a well defined constant $a$) in order to achieve the best convergence rate.

\subsection{Lower Bound Convergence Rate}



Applying the lower bound from \citep{nguyen2019tight} for first order stochastic algorithms shows the following corollary, see Supplemental Material \ref{sec:tight} for a detailed discussion (also on how fast the $O((\ln t) /t)$ term disappears and how this can be influenced by using different less increasing sample size sequences).

\begin{cor} \label{lem:lower} Among first order stochastic algorithms, 
upper bound (\ref{eqUCR}) 
converges for increasing $t$ to within 
a constant factor $8\cdot 36^2$ of the (theoretically) best attainable expected convergence rate, which is at least $ \frac{N}{2\mu^2} \frac{1}{t} (1- O(\frac{\ln t}{t}))$ (for each $t$). 
\end{cor}

Notice that the factor is independent of any parameters like $L$, $\mu$, sparsity, or dimension of the model.

The corollary shows that non-parallel (and therefore synchronous) SGD can at most achieve a factor $8\cdot 36^2$ faster convergence rate compared to our asynchronous SGD over (heterogeneous) local data sets. It remains an open problem to investigate second (and higher) order stochastic algorithms and whether their distributed versions attain tight convergence rates when increasing sample sizes and diminishing step sizes from round to round.

\section{Experiments}
\label{sec:experiment}
We summarize experimental results for strongly convex, plain convex and non-convex problems with \textit{linear increasing sample sequences} and \textit{biased versus unbiased local data sets}.

As the plain convex objective function we use logistic regression: The weight vector $\bar{w}$ and bias value $b$ of the logistic function can be learned by minimizing the log-likelihood function $J$:
\begin{equation} 
\nonumber
    J = - \sum_{i}^M [ y_{i} \cdot \log (\sigma_i) + (1 - y_{i}) \cdot \log (1 - \sigma_i) ],
\end{equation}
where $M$ is the number of training samples $(x_i,y_i)$ with $y_i\in\{0,1\}$, and $\sigma_i=\sigma(\bar{w},b,x_i)$ is the sigmoid function 
$    \sigma(\bar{w}, x, b) = \frac{1}{1 + e^{-(\bar{w}^{\mathrm{T}}x + b)}}
$.
The goal is to learn a vector $w^*$ which represents a pair $w=(\bar{w},b)$ that minimizes $J$.
Function $J$  changes into a strongly convex problem by adding ridge regularization with a regularization parameter $\lambda>0$. i.e., we minimize $\hat{J}=J+ \frac{\lambda}{2} \norm{w}^2$ instead of $J$.
For simulating non-convex problems, we choose a simple neural network (LeNet) \citep{lecun1998gradient} for image classification.

We use a linearly increasing sample size sequence $s_i = a \cdot i^c + b$, where $c=1$ and $a,b \geq 0$. For simplicity, we choose a diminishing round step size sequence corresponding to $\frac{\eta_0}{1 + \beta \cdot t}$ for the strongly convex problem and $\frac{\eta_0}{1 + \beta \cdot \sqrt{t}}$ for both the plain convex and non-convex problems, where $\eta_0$ is an initial step size. The asynchronous SGD simulation is conducted with $d=1$, see (\ref{eqtausample}).

\begin{figure}[ht!]
\vspace{-.50cm}
  \centering
  \subfloat[Convergence rate]{\includegraphics[width=0.50\textwidth]{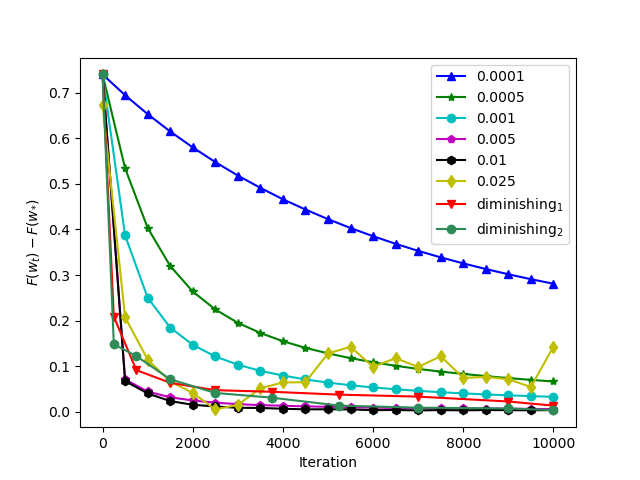}\label{fig:async_fl_strongly_convex_phishing_convergence_main}}
  \hfill
  \subfloat[Test error]{\includegraphics[width=0.50\textwidth]{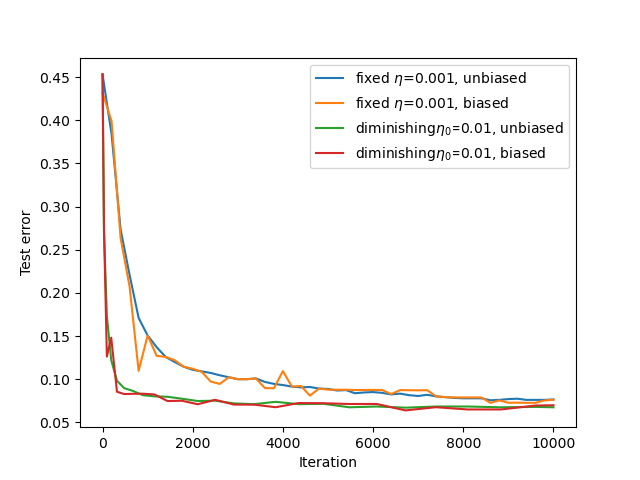}\label{fig:asyn_fl_biased_strongly_convex_mnist_main}}
  \caption{Our asynchronous SGD for strongly convex problems: (a) The Phishing data set - various step size sequences. (b) MNIST - biased and unbiased data sets.}
  \label{fig:async_strongly_convex_phishing_main}
\end{figure}
\begin{figure}[ht!]
\vspace{-.40cm}
  \centering
  \subfloat[Convergence rate]{\includegraphics[width=0.50\textwidth]{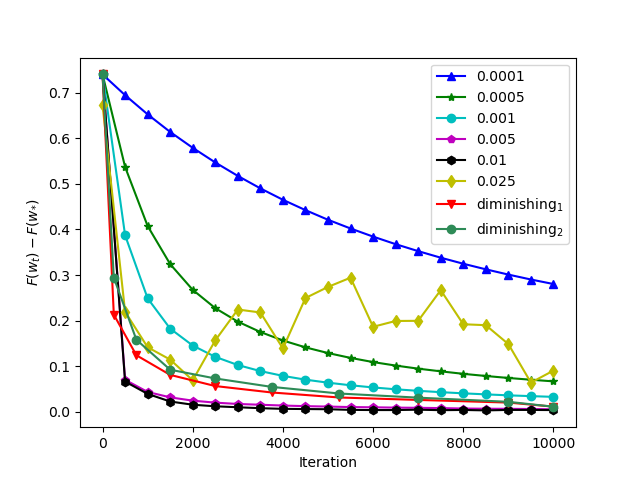}\label{fig:async_fl_plain_convex_phishing_convergence_main}}
  \hfill
  \subfloat[Test error]{\includegraphics[width=0.50\textwidth]{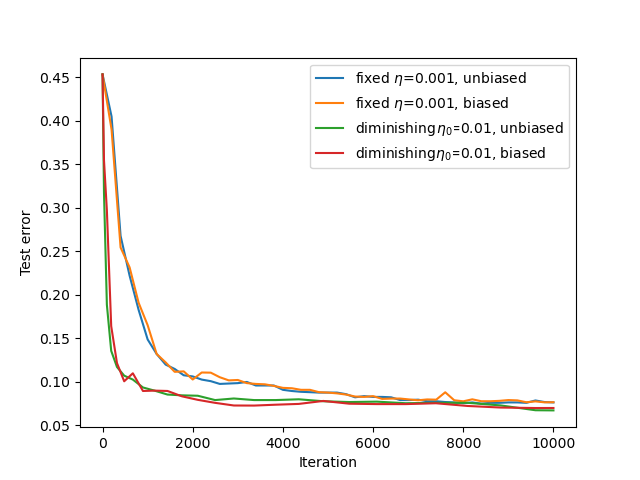}\label{fig:asyn_fl_biased_plain_convex_mnist_main}}
  \caption{
  Our asynchronous SGD for plain convex problems: (a) The Phishing data set - various step size sequences. (b) MNIST - biased and unbiased data set.}
  \label{fig:async_plain_convex_phishing_main}
\end{figure}
\begin{figure}[H]
\vspace{-.50cm}
  \centering
  \subfloat[Test error]{\includegraphics[width=0.50\textwidth]{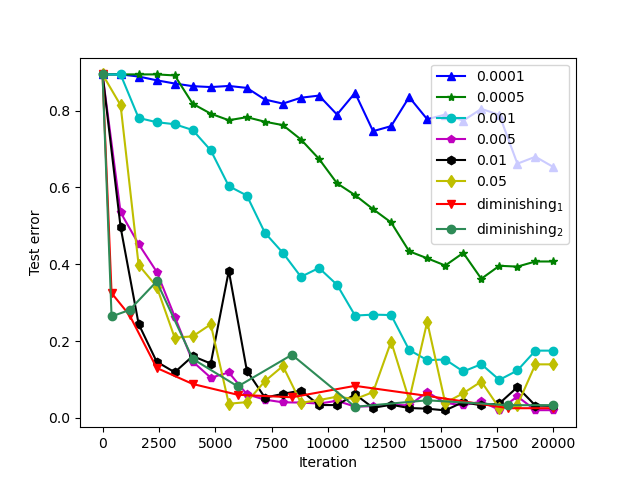}\label{fig:async_fl_non_convex_mnist_err_main}}
  \hfill
  \subfloat[Test error]{\includegraphics[width=0.50\textwidth]{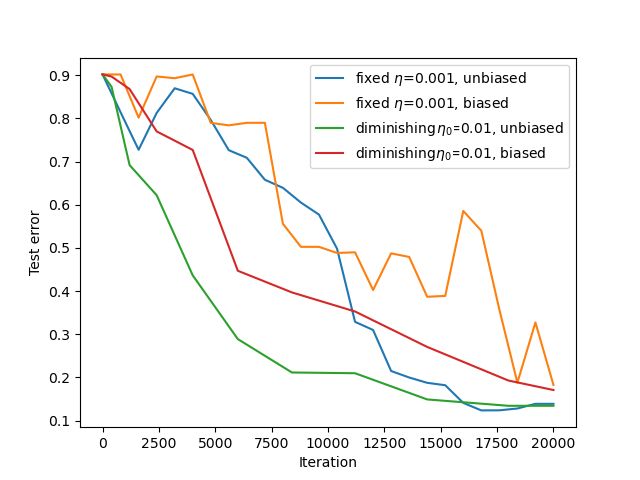}\label{fig:asyn_fl_biased_non_convex_mnist_main}}
  \caption{
  Our asynchronous SGD for non-convex problems: (a) The MNIST data set - various step size sequences. (b) MNIST - biased and unbiased data set.}
  \label{fig:async_non_convex_mnist_main}
\end{figure}

\textbf{Our asynchronous SGD with linear increasing sample sizes:} 
Figures~\ref{fig:async_fl_strongly_convex_phishing_convergence_main}, \ref{fig:async_fl_plain_convex_phishing_convergence_main} and ~\ref{fig:async_fl_non_convex_mnist_err_main} show our proposed asynchronous SGD with linear increasing sample size sequence for constant step sizes and diminishing step sizes. We conclude that diminishing step sizes achieve (approximately) a convergence which is as fast as the convergence of the best constant step size sequence. 
See Supplemental Material
~\ref{subsec:stepsize_FL_simu}
for details and larger sized graphs (also for other data sets); we also show experiments in~\ref{subsec:constant_stepsize_samplesize_FL_simu} and ~\ref{subsec:samplesize_FL_simu}  that compare using slower increasing sample size sequences with linear sample size sequences and they all achieve approximately the same convergence rate. We conclude that using a linear sample size sequence over a constant sized one does not degrade performance -- on the contrary, the number of communication rounds reduces significantly. This confirms the intuition generated by our theoretical analysis for strong convex problems which generalizes to plain and non-convex problems.


\textbf{Our asynchronous SGD with  biased data sets:} 
The goal of this experiment is to show that our asynchronous SGD can work well with biased (non-iid) local data sets meaning that different compute nodes use different distributions ${\cal D}_c$ (contrary to all nodes using unbiased data sets such that they all have the same ${\cal D}_c={\cal D}$ distribution). 
We continue with the setting as mentioned above with an adapted initial step size $\eta_{0}$, see Supplemental Material~\ref{subsec:dataset_FL_simu} for details.
Figures~\ref{fig:asyn_fl_biased_strongly_convex_mnist_main} and \ref{fig:asyn_fl_biased_plain_convex_mnist_main} show no significant difference 
between using biased or unbiased data sets for strongly convex and plain convex problems.
For the non-convex problem, Figure~\ref{fig:asyn_fl_biased_non_convex_mnist_main} shows that although the accuracy might fluctuate during the training process, our asynchronous SGD still achieves good accuracy in general. We conclude that our asynchronous SGD 
tolerates the effect of biased data sets, which is quite common in practice.

\textbf{Scalability:} Supplemental Material \ref{subsec:num_worker} shows that our asynchronous SGD with linear sample size sequence  scales to larger number of compute nodes. The accuracy for the same total number of gradient computations stays approximately the same and the overall execution time will reach a lower limit (where increased parallelism does not help). The linear sample size sequence gives a reduced number of communication rounds (also for an increased number of compute nodes).


\section{Conclusion}
We provided a tight theoretical analysis for strongly convex problems over heterogeneous local data for our asynchronous SGD with increasing sample size sequences. Experiments confirm that not only strongly convex but also plain and non-convex problems can tolerate linear increasing sample sizes -- this reduces the number of communication rounds.

\bibliography{references}

\begin{thebibliography}{35}
\providecommand{\natexlab}[1]{#1}
\providecommand{\url}[1]{\texttt{#1}}
\expandafter\ifx\csname urlstyle\endcsname\relax
  \providecommand{\doi}[1]{doi: #1}\else
  \providecommand{\doi}{doi: \begingroup \urlstyle{rm}\Url}\fi

\bibitem[Bonawitz et~al.(2019)Bonawitz, Eichner, Grieskamp, Huba, Ingerman,
  Ivanov, Kiddon, Konecny, Mazzocchi, McMahan, et~al.]{bonawitz2019towards}
Keith Bonawitz, Hubert Eichner, Wolfgang Grieskamp, Dzmitry Huba, Alex
  Ingerman, Vladimir Ivanov, Chloe Kiddon, Jakub Konecny, Stefano Mazzocchi,
  H~Brendan McMahan, et~al.
\newblock Towards federated learning at scale: System design.
\newblock \emph{arXiv preprint arXiv:1902.01046}, 2019.

\bibitem[Bottou et~al.(2018)Bottou, Curtis, and Nocedal]{BottouCN18}
L{\'{e}}on Bottou, Frank~E. Curtis, and Jorge Nocedal.
\newblock Optimization methods for large-scale machine learning.
\newblock \emph{{SIAM} Review}, 60\penalty0 (2):\penalty0 223--311, 2018.

\bibitem[Chee and Toulis(2018)]{CheeT18}
Jerry Chee and Panos Toulis.
\newblock Convergence diagnostics for stochastic gradient descent with constant
  learning rate.
\newblock In \emph{International Conference on Artificial Intelligence and
  Statistics, {AISTATS} 2018}, pages 1476--1485, 2018.

\bibitem[Chen et~al.(2016)Chen, Monga, Bengio, and Jozefowicz]{jianmin}
Jianmin Chen, Rajat Monga, Samy Bengio, and Rafal Jozefowicz.
\newblock Revisiting distributed synchronous sgd.
\newblock \emph{ICLR Workshop Track}, 2016.

\bibitem[Chen et~al.(2019{\natexlab{a}})Chen, Sun, and Jin]{yang}
Yang Chen, Xiaoyan Sun, and Yaochu Jin.
\newblock Communication-efficient federated deep learning with asynchronous
  model update and temporally weighted aggregation.
\newblock \emph{arXiv preprint}, 2019{\natexlab{a}}.
\newblock URL \url{https://arxiv.org/pdf/1903.07424.pdf}.

\bibitem[Chen et~al.(2019{\natexlab{b}})Chen, Sun, and Jin]{yang2019}
Yang Chen, Xiaoyan Sun, and Yaochu Jin.
\newblock Communication-efficient federated deep learning with asynchronous
  model update and temporally weighted aggregation.
\newblock \emph{arXiv preprint}, 2019{\natexlab{b}}.
\newblock URL \url{https://arxiv.org/pdf/1903.07424.pdf}.

\bibitem[De~Sa et~al.(2015)De~Sa, Zhang, Olukotun, and
  R{\'e}]{DeSaZhangOlukotunEtAl2015}
Christopher~M De~Sa, Ce~Zhang, Kunle Olukotun, and Christopher R{\'e}.
\newblock {Taming the wild: A unified analysis of hogwild-style algorithms}.
\newblock In \emph{NIPS}, pages 2674--2682, 2015.

\bibitem[Hsieh et~al.(2017)Hsieh, Harlap, Vijaykumar, Konomis, Ganger, Gibbons,
  and Mutlu]{kevin}
Kevin Hsieh, Aaron Harlap, Nandita Vijaykumar, Dimitris Konomis, Gregory~R.
  Ganger, Phillip~B. Gibbons, and Onur Mutlu.
\newblock Gaia: Geo-distributed machine learning approaching {LAN} speeds.
\newblock \emph{14th {USENIX} Symposium on Networked Systems Design and
  Implementation ({NSDI} 17).}, 2017.

\bibitem[Jie~Xu(2020)]{jie2020asynchronous}
Fei~Wang Jie~Xu, Wei~Zhang.
\newblock Asynchronous decentralized parallel stochastic gradient descent with
  differential privacy.
\newblock \emph{arXiv preprint arXiv:2008.09246}, 2020.

\bibitem[Khaled et~al.(2020)Khaled, Mishchenko, and
  Richt{\'a}rik]{khaled2020tighter}
Ahmed Khaled, Konstantin Mishchenko, and Peter Richt{\'a}rik.
\newblock Tighter theory for local sgd on identical and heterogeneous data.
\newblock In \emph{International Conference on Artificial Intelligence and
  Statistics}, pages 4519--4529. PMLR, 2020.

\bibitem[Kone{\v{c}}n{\`y} et~al.(2016)Kone{\v{c}}n{\`y}, McMahan, Ramage, and
  Richt{\'a}rik]{konevcny2016federated}
Jakub Kone{\v{c}}n{\`y}, H~Brendan McMahan, Daniel Ramage, and Peter
  Richt{\'a}rik.
\newblock Federated optimization: Distributed machine learning for on-device
  intelligence.
\newblock \emph{arXiv preprint arXiv:1610.02527}, 2016.

\bibitem[Leblond et~al.(2018)Leblond, Pedregosa, and
  Lacoste-Julien]{Leblond2018}
R{\'e}mi Leblond, Fabian Pedregosa, and Simon Lacoste-Julien.
\newblock Improved asynchronous parallel optimization analysis for stochastic
  incremental methods.
\newblock \emph{JMLR}, 19\penalty0 (1):\penalty0 3140--3207, 2018.

\bibitem[LeCun et~al.(1998)LeCun, Bottou, Bengio, and
  Haffner]{lecun1998gradient}
Yann LeCun, L{\'e}on Bottou, Yoshua Bengio, and Patrick Haffner.
\newblock Gradient-based learning applied to document recognition.
\newblock \emph{Proceedings of the IEEE}, 86\penalty0 (11):\penalty0
  2278--2324, 1998.

\bibitem[Li et~al.(2019)Li, Sahu, Zaheer, Sanjabi, Talwalkar, and
  Smith]{tianli2019hetero}
Tian Li, Anit~Kumar Sahu, Manzil Zaheer, Maziar Sanjabi, Ameet Talwalkar, and
  Virginia Smith.
\newblock Federated optimization for heterogeneous networks.
\newblock \emph{arXiv preprint}, 2019.
\newblock URL \url{https://arxiv.org/pdf/1812.06127.pdf}.

\bibitem[Lian et~al.(2015)Lian, Huang, Li, and Liu]{lian2015asynchronous}
Xiangru Lian, Yijun Huang, Yuncheng Li, and Ji~Liu.
\newblock Asynchronous parallel stochastic gradient for nonconvex optimization.
\newblock In \emph{Advances in Neural Information Processing Systems}, pages
  2737--2745, 2015.

\bibitem[Lian et~al.(2017)Lian, Zhang, Zhang, and Liu]{lian2017asynchronous}
Xiangru Lian, Wei Zhang, Ce~Zhang, and Ji~Liu.
\newblock Asynchronous decentralized parallel stochastic gradient descent.
\newblock \emph{arXiv preprint arXiv:1710.06952}, 2017.

\bibitem[Mania et~al.(2015)Mania, Pan, Papailiopoulos, Recht, Ramchandran, and
  Jordan]{ManiaPanPapailiopoulosEtAl2015}
Horia Mania, Xinghao Pan, Dimitris Papailiopoulos, Benjamin Recht, Kannan
  Ramchandran, and Michael~I Jordan.
\newblock {Perturbed Iterate Analysis for Asynchronous Stochastic
  Optimization}.
\newblock \emph{SIAM Journal on Optimization}, pages 2202--2229, 2015.

\bibitem[McMahan and Ramage(2017)]{GoogleAIBlog}
Brendan McMahan and Daniel Ramage.
\newblock Federated learning: Collaborative machine learning without
  centralized training data, 2017.
\newblock URL
  \url{https://ai.googleblog.com/2017/04/federated-learning-collaborative.html}.
\newblock Last accessed 09/24/2019.

\bibitem[McMahan et~al.(2016)McMahan, Moore, Ramage, and y~Arcas]{mcmahan}
H.~Brendan McMahan, Eider Moore, Daniel Ramage, and Blaise~Agüera y~Arcas.
\newblock Federated learning of deep networks using model averaging.
\newblock \emph{ICLR Workshop Track}, 2016.

\bibitem[Meng et~al.(2017)Meng, Chen, Yu, Wang, Ma, and
  Liu]{meng2017asynchronous}
Qi~Meng, Wei Chen, Jingcheng Yu, Taifeng Wang, Zhi-Ming Ma, and Tie-Yan Liu.
\newblock Asynchronous stochastic proximal optimization algorithms with
  variance reduction.
\newblock In \emph{Thirty-First AAAI Conference on Artificial Intelligence},
  2017.

\bibitem[Nguyen et~al.(2018)Nguyen, Nguyen, van Dijk, Richt{\'{a}}rik,
  Scheinberg, and Tak{\'{a}}c]{nguyen2018sgd}
Lam~M. Nguyen, Phuong~Ha Nguyen, Marten van Dijk, Peter Richt{\'{a}}rik, Katya
  Scheinberg, and Martin Tak{\'{a}}c.
\newblock {SGD} and hogwild! convergence without the bounded gradients
  assumption.
\newblock In \emph{Proceedings of the 35th International Conference on Machine
  Learning, {ICML} 2018}, pages 3747--3755, 2018.

\bibitem[Nguyen et~al.(2019{\natexlab{a}})Nguyen, Nguyen, Richt{{\'a}}rik,
  Scheinberg, Tak{{\'a}}{\v{c}}, and van Dijk]{nguyen2018new}
Lam~M. Nguyen, Phuong~Ha Nguyen, Peter Richt{{\'a}}rik, Katya Scheinberg,
  Martin Tak{{\'a}}{\v{c}}, and Marten van Dijk.
\newblock New convergence aspects of stochastic gradient algorithms.
\newblock \emph{Journal of Machine Learning Research}, 20\penalty0
  (176):\penalty0 1--49, 2019{\natexlab{a}}.

\bibitem[Nguyen et~al.(2019{\natexlab{b}})Nguyen, Nguyen, and van
  Dijk]{nguyen2019tight}
P.~H. Nguyen, L.~M. Nguyen, and M.~van Dijk.
\newblock Tight dimension independent lower bound on the expected convergence
  rate for diminishing step sizes in {SGD}.
\newblock \emph{The 33th Annual Conference on Neural Information Processing
  Systems (NeurIPS 2019)}, 2019{\natexlab{b}}.

\bibitem[Recht et~al.(2011)Recht, Re, Wright, and Niu]{Hogwild}
Benjamin Recht, Christopher Re, Stephen Wright, and Feng Niu.
\newblock Hogwild: A lock-free approach to parallelizing stochastic gradient
  descent.
\newblock In \emph{Advances in neural information processing systems}, pages
  693--701, 2011.

\bibitem[Robbins and Monro(1951)]{RM1951}
Herbert Robbins and Sutton Monro.
\newblock A stochastic approximation method.
\newblock \emph{The Annals of Mathematical Statistics}, 22\penalty0
  (3):\penalty0 400--407, 1951.

\bibitem[Saeed~Ghadimi(2013)]{Ghadimi2013Minibatch}
Hongchao~Zhang Saeed~Ghadimi, Guanghui~Lan.
\newblock Mini-batch stochastic approximation methods for nonconvex stochastic
  composite optimization.
\newblock \emph{arXiv preprint arxiv:1308.6594}, 2013.

\bibitem[Shi et~al.(2019)Shi, Wang, Zhao, Tang, Wang, Huang, and
  Chu]{shi2019distributed}
Shaohuai Shi, Qiang Wang, Kaiyong Zhao, Zhenheng Tang, Yuxin Wang, Xiang Huang,
  and Xiaowen Chu.
\newblock A distributed synchronous sgd algorithm with global top-$ k $
  sparsification for low bandwidth networks.
\newblock \emph{arXiv preprint arXiv:1901.04359}, 2019.

\bibitem[Spiridonoff et~al.(2020)Spiridonoff, Olshevsky, and
  Paschalidis]{spiridonoff2020local}
Artin Spiridonoff, Alex Olshevsky, and Ioannis~Ch Paschalidis.
\newblock Local sgd with a communication overhead depending only on the number
  of workers.
\newblock \emph{arXiv preprint arXiv:2006.02582}, 2020.

\bibitem[Stich(2018)]{stich2018local}
Sebastian~U Stich.
\newblock Local sgd converges fast and communicates little.
\newblock \emph{arXiv preprint arXiv:1805.09767}, 2018.

\bibitem[Van~Dijk et~al.(2019)Van~Dijk, Nguyen, Nguyen, and
  Phan]{van2019characterization}
Marten Van~Dijk, Lam Nguyen, Phuong~Ha Nguyen, and Dzung Phan.
\newblock Characterization of convex objective functions and optimal expected
  convergence rates for sgd.
\newblock In \emph{International Conference on Machine Learning}, pages
  6392--6400, 2019.

\bibitem[Wang et~al.(2019)Wang, Wang, and Li]{luping}
Luping Wang, Wei Wang, and Bo~Li.
\newblock Cmfl: Mitigating communication overhead for federated learning.
\newblock \emph{IEEE International Conference on Distributed Computing
  Systems.}, 2019.

\bibitem[Xie et~al.(2019)Xie, Koyejo, and Gupta]{cong2019}
Cong Xie, Sanmi Koyejo, and Indranil Gupta.
\newblock Asynchronous federated optimization.
\newblock \emph{arXiv preprint}, 2019.
\newblock URL \url{https://arxiv.org/pdf/1903.03934v1.pdf}.

\bibitem[Yu and Jin(2019)]{yu2019computation}
Hao Yu and Rong Jin.
\newblock On the computation and communication complexity of parallel sgd with
  dynamic batch sizes for stochastic non-convex optimization.
\newblock In \emph{International Conference on Machine Learning}, pages
  7174--7183. PMLR, 2019.

\bibitem[Zheng et~al.(2017)Zheng, Meng, Wang, Chen, Yu, Ma, and
  Liu]{zheng2017asynchronous}
Shuxin Zheng, Qi~Meng, Taifeng Wang, Wei Chen, Nenghai Yu, Zhi-Ming Ma, and
  Tie-Yan Liu.
\newblock Asynchronous stochastic gradient descent with delay compensation.
\newblock In \emph{Proceedings of the 34th International Conference on Machine
  Learning-Volume 70}, pages 4120--4129. JMLR. org, 2017.

\bibitem[Zinkevich et~al.(2009)Zinkevich, Langford, and
  Smola]{zinkevich2009slow}
Martin Zinkevich, John Langford, and Alex~J Smola.
\newblock Slow learners are fast.
\newblock In \emph{Advances in neural information processing systems}, pages
  2331--2339, 2009.

\end{thebibliography}
\bibliographystyle{plainnat}

\clearpage

\appendix

\vbox{%
    \hsize\textwidth
    \linewidth\hsize
    \vskip 0.1in
    \hrule height 4pt
    \vskip 0.25in
    \vskip -5.5pt%
    \centering
    {\Large\bf{Supplementary Material} \par}
    \vskip 0.29in
    \vskip -5.5pt
    \hrule height 1pt
    \vskip 0.09in%
    \vskip 0.2in
}
  

\section{Algorithms} \label{app-alg}

\begin{algorithm}[!ht]
\caption{Initial Setup} 
\label{alg:setup}

\begin{algorithmic}[1]
\Procedure{Setup}{$n$}
    \State {\bf Initialize} global model $\hat{v}_0=\hat{w}_{c,0,0}$ for server and compute nodes $c\in \{1,\ldots, n\}$
    \State {\bf Initialize} diminishing round step size sequence $\{\bar{\eta}_{i}\}_{i\geq 0}$
    \State {\bf Initialize} increasing sample size sequence $\{s_{i}\}_{i\geq 0}$
    \For{$i \geq 0$}
        \For{$t \in \{0,\ldots, s_i-1\}$}
            \State Assign $a(i,t)=c$ with probability $p_c$
        \EndFor
        \For{$c\in \{1,\ldots n\}$}
            \State $s_{i,c}= |\{ t \ : \ a(i,t)=c\}|$
        \EndFor
    \EndFor
    \State \Comment{$\{s_{i,c}\}_{i\geq 0}$ represents the sample size sequence for compute node $c$; notice that $\mathbb{E}[s_{i,c}]=p_c s_i$}
    \State {\bf Initialize} permissible delay function $\tau(\cdot)$ with $t-\tau(t)$ increasing in $t$
\EndProcedure
\end{algorithmic}
\end{algorithm}

\begin{algorithm}[htb]
\caption{Server -- Global Model}
\label{alg:server}
 
\begin{algorithmic}[1]
\Procedure{ISRReceive}{$message$} \Comment{Interrupt Service Routine}
\If{$message==(i,c,U)$ is from a compute node $c$} \Comment{$U$ represents $U_{i,c}$}
\State $Q.enqueue(message)$ 
 \Comment{Queue $Q$ maintains aggregate gradients not yet processed}
\EndIf
\EndProcedure
\State
\Procedure{MainServer}{}
\State $k = 0$ \Comment{Represents a broadcast counter}
\State $\hat{v}=\hat{v}_0$
\State Initialize $Q$ and $H$ to empty queues
\While{\textbf{True}}
       
    \If{$Q$ is not empty}
    \State {$(i,c,U) \leftarrow$ $Q.dequeue()$} \Comment{Receive $U_{i,c}$}
            \State $\hat{v} = \hat{v} - \bar{\eta}_{i} \cdot U$
            \State $H.enqueue((i,c))$  
            \If{$H$ has $(k,c)$ for all $c\in \{1,\ldots, n\}$}
                \State $H.dequeue((k,c))$ for all $c \in \{1,\ldots, n\}$ 
             \State $k$++
                \State Broadcast $(\hat{v}, k)$ to all compute nodes \Comment{$\hat{v}=\hat{v}_k$}
            \EndIf
    \EndIf
    \State \Comment{{\em Invariant}: $\hat{v}= \hat{v}_0 -
    \sum_{i=0}^{k-1} \sum_{c=1}^n \bar{\eta}_i U_{i,c}
    - \sum_{(i,c)\in H} \bar{\eta}_i U_{i,c}$}
    \State \Comment{$\hat{v}_k$ 
    includes the aggregate of updates $\sum_{i=0}^{k-1} \sum_{c=1}^n \bar{\eta}_i U_{i,c}$}

\EndWhile
\EndProcedure

\end{algorithmic}
\end{algorithm}

\begin{algorithm}[!tb]
\caption{ComputeNode$_c$ -- Local Model}
\label{alg:client}

\begin{algorithmic}[1]

\Procedure{ISRReceive}{$message$} \Comment{Interrupt Service Routine}
\If{$message==(\hat{v}, k)$ comes from the server \& $k > kold$} 
\State \Comment{The compute node will only accept and use a more fresh global model}
\State Replace the variable $k$ as maintained by the compute node by that of  $message$

            \State $\hat{w} = \hat{v} - \bar{\eta}_i \cdot U$ \Comment{This represents $\hat{w}_{c,i,h}= \hat{v}_k - \bar{\eta}_i \cdot U_{i,c,h}$}

\State $kold=k$
\State $t_{delay} = s_{k}+\ldots+s_{i} - (s_{i,c}-h) $
\EndIf
\EndProcedure


\State
\Procedure{MainComputeNode}{$\mathcal{D}_c$} \Comment{$\mathcal{D}_c$ represents the local training data for compute node $c$}
\State $i=0$, $\hat{w}=\hat{w}_{c,0,0}$ \Comment{Local round counter}

\While{\textbf{True}}
    \State $h=0$, $U=0$ \Comment{$U_{i,c,h}$ for $h=0$ equals the all-zero vector}
    \While{$h< s_{i,c}$}
       \State $t_{glob}= s_0+\ldots+s_{i} - (s_{i,c}-h)-1$ 
       \State $t_{delay} = s_{k}+\ldots+s_{i} - (s_{i,c}-h) $
 
        \While{$\tau(t_{glob}) <  t_{delay}$}
            \State only keep track of the update of $t_{delay}$
            \Comment{{\em Invariant}: $t_{delay}\leq \tau(t_{glob})$}
        \EndWhile
        
        \State Sample uniformly at random $\xi$ from $\mathcal{D}_c$ \Comment{$\xi$ represents $\xi_{c,i,h}$}
        
        \State $g = \nabla f(\hat{w}, \xi)$ \Comment{$g$ represents $g_{c,i,h}= \nabla f(\hat{w}_{c,i,h}, \xi_{c,i,h}) $}
        
        \State ${U} = {U} + g$  \Comment{Represents $U_{i,c,h+1}=U_{i,c,h} + g_{c,i,h}$ implying $U_{i,c,h+1}= \sum_{j=0}^h g_{c,i,j}$}
        
        \State Update model $w = \hat{w} - \bar{\eta}_{i} \cdot g$ \Comment{This represents $ w_{c,i,h+1} = \hat{w}_{c,i,h} - \bar{\eta}_i g_{c,i,h}$}
         \State Update model $\hat{w} = w$ \Comment{This represents $\hat{w}_{c,i,h+1} = w_{c,i,h+1}$}
        
        \State $h$++
    \EndWhile
    
    \State Send $(i,c, U)$ to the Server. \Comment{$U$ represents $U_{i,c} = U_{i,c,s_{i,c}} = \sum_{h=0}^{s_{i,c}-1} g_{c,i,h}$}
    \State $i$++
\EndWhile
   
\EndProcedure

\end{algorithmic}
\end{algorithm} 


The centralized aggregation server maintains a global model which is updated according to Algorithm \ref{alg:server}.
The server receives updates from compute nodes who work on local models by executing Algorithm \ref{alg:client}. Before any computation starts both server and clients agree on the used diminishing step size sequence, the increasing sample (mini-batch) sequences, initial default global model, and permissible delay function $\tau$, see Algorithm \ref{alg:setup}.
Note that the Interrupt Service Routines (ISR) at the compute node and server  interrupt the main execution as soon as a message is received in which case the ISR routines execute before returning to the main code.

The algorithms have added comments with interpretation of the locally computed variables. These interpretations/annotations are used in next sections when proving properties.

We remark that $s_{i,c}$ are initialized by a coin flipping procedure in \Call{Setup}{}. Since the $s_i$ are increasing, we may approximate $s_{i,c}\approx \mathbb{E}[s_{i,c}]=p_c s_i$ (because of the law of large numbers). 

\section{Proofs asynchronous distributed SGD} 

\subsection{Proof of Theorem \ref{thmalg}}
\label{appmain}

The clients in the distributed computation apply recursion (\ref{eqwM2a}). We want to label each recursion with an iteration count $t$; this can then be used to compute with which delay function the labeled sequence $\{w_t \}$ is consistent.
In order to find an ordering based on $t$ we first define a mapping $\rho$ from the annotated labels $(c,i,h)$ in \Call{MainComputeNode}{} to $t$:
%
$$\rho(c,i,h)= (\sum_{l < i} s_{l}) + \min\{ t' \ : \ h=|\{ t\leq t' \ : \ a(i,t)=c\}|\} ,$$ 
where sequence sample size sequence $\{s_i\}$ and labelling function $a(\cdot,\cdot)$ are defined in \Call{Setup}{}.

Notice that given $t$, we can compute $i$ as the largest index for which $\sum_{l<i} s_{l} \leq t$, compute $t'=t- \sum_{l<i} s_{l}$ and $c=a(i,t')$, and compute $h=|\{ t\leq t' \ : \ a(i,t)=c\}|$. This procedure inverts $\rho$ since $\rho(c,i,h)=t$, hence, $\rho$ is bijective.  We use $\rho$ to order local models $\hat{w}_{c,i,h}$ by writing $\hat{w}_t=\hat{w}_{c,i,h}$ with $t=\rho(c,i,h)$. Similarly, we write $\xi_t= \xi_{c,i,h}$ for local training data samples. 


From the invariant in \Call{MainServer}{} we infer that  $\hat{v}_k$ includes all the aggregate updates $U_{i,c}$ for $i< k$ and $c\in {1,\ldots n}$. See Algorithm \ref{alg:client} for  compute nodes, 
$$\sum_{c\in \{1,\ldots, n\}} U_{i,c} = \sum_{c\in \{1, \ldots, n\}} \sum_{h=0}^{s_{i,c}-1} g_{c,i,h} \mbox{ with } g_{c,i,h}= \nabla f(\hat{w}_{c,i,h}; \xi_{c,i,h}).$$
By using mapping $\rho$, this is equal to
\begin{eqnarray*}
\sum_{c\in \{1, \ldots, n\}} \sum_{h=0}^{s_{i,c}-1} \nabla f(\hat{w}_{c,i,h}; \xi_{c,i,h}) &=&
\sum_{c\in \{1, \ldots, n\}} \sum_{h=0}^{s_{i,c}-1} \nabla f(\hat{w}_{\rho(c,i,h)}; \xi_{\rho(c,i,h)}) \\
&=& \sum_{t=s_0+\ldots s_{i-1}}^{s_0+\ldots+s_i-1} \nabla f(\hat{w}_{t}; \xi_{t}).
\end{eqnarray*}
This implies that 
$\hat{v}_k$, includes all the gradient updates (across all clients) that correspond to $\hat{w}_t$ for $t\leq s_0+\ldots+s_{k-1}-1= t_{glob}-t_{delay}$ in the notation of \Call{MainComputeNode}{}.

Let $\rho(c,i,h)=t$. Notice that $t= \rho(c,i,h) \leq s_0+ \ldots + s_i -(s_{i,c}-h)-1=t_{glob}$.
In \Call{MainComputeNode}{} we wait as long as $\tau(t_{glob})= t_{delay}$. This means that $h$ will not further increase until $k$ increases in compute node $c$'s \Call{ISRReceive}{} when a new global model is received. This implies $t_{delay}\leq \tau(t_{glob})$ as an invariant of the algorithm. Because $t-\tau(t)$ is an increasing function in $t$, the derived inequality implies  $t-\tau(t)\leq t_{glob}-\tau(t_{glob}) \leq t_{glob}-t_{delay}$. Notice that $\hat{w}_t=\hat{w}_{c,i,h}$  includes the updates aggregated in $\hat{v}_k$ as last received by compute node $c$'s \Call{ISRReceive}{}. We concluded above that $\hat{v}_k$ includes all gradient updates that correspond to gradient computations up to iteration count $t_{glob}-t_{delay}$. This includes gradient computations up to iteration count $t-\tau(t)$. Therefore, the computed sequence $\{w_t\}$ is consistent with delay function $\tau$.

As a second consequence of mapping $\rho$ we notice that $\{\xi_t\}=\{\xi_{c,i,h}\}$. In fact $\xi_t=\xi_{c,i,h}$ for $(c,i,h)=\rho^{-1}(t)$. 
Notice that \Call{Setup}{} constructs the mapping $a(i,t)=c$  which is used to define $\rho$. Mapping $a(i,t)$ represents a random table of compute node assignments $c$ generated by using a probability vector $(p_1,\ldots, p_n)$ (with $c$ being selected with probability $p_c$ for the $(i,t)$-th entry). So, $\xi_t\sim {\cal D}_c$ in \Call{MainClient}{} with probability $p_c$. This means that $\xi_t\sim {\cal D}$ with ${\cal D} = \sum_{c=1}^n p_c {\cal D}_c$.

\subsection{Invariant $t_{delay}\leq \tau(t_{glob})$} \label{appinv}

We propose to use increasing sample size sequences $\{s_{i,c}\}$ such that we can replace the condition $\tau(t_{glob})<t_{delay}$ 
of the wait loop by 
$i > k + d$, 
where $d$ is a threshold such that for all $i\geq d$,
\begin{equation} \tau(\sum_{j=0}^i s_j)\geq 1+\sum_{j=i-d}^i s_j.
\label{eqS}
\end{equation}
The new wait loop guarantees that $i-k\leq d$ as an invariant of the algorithm. This implies the old invariant $t_{delay}\leq \tau(t_{glob})$ for the following reason: Let $i\geq d$. Since $t-\tau(t)$ is increasing,
$$t_{glob}-\tau(t_{glob}) \leq \sum_{j=0}^i s_j -\tau(\sum_{j=0}^i s_j) \leq \sum_{j=0}^i s_j - (1+\sum_{j=i-d}^i s_j) =-1+ \sum_{j=0}^{i-d-1} s_i.$$
Together with $i-d\leq k$, this implies
$$ \tau(t_{glob}) \geq t_{glob} -(-1+ \sum_{j=0}^{i-d-1} s_i) = s_{i-d}+\ldots + s_i -(s_{i,c}-h)  \geq 
s_{k}+\ldots + s_i -(s_{i,c}-h)= t_{delay}.$$
This shows that the new invariant implies the old one when $i\geq d$.

\subsection{Increasing sample size sequences} \label{appinc}

The following lemma shows how to construct an increasing sample size sequence given a delay function $\tau$. 

\begin{lem} \label{lemsample}
Let $g>1$.
Let function $\gamma(z)$  be increasing (i.e., $\gamma'(z)\geq 0$)  and $\geq 1$ for $z\geq 0$, with the additional property that 
$\gamma(z)\geq z \gamma'(z) \frac{g}{g-1}$
for $z\geq 0$. Suppose that 
$$ \tau(x) = M_1 +   (\frac{x+M_0}{\gamma(x+M_0)})^{1/g}  
\mbox{ with } 
M_0 \geq ((m+1)\frac{g-1}{g})^{g/(g-1)} \mbox{ and } M_1\geq d+2 $$
for some integer $m\geq 0$
(we can choose $M_0=0$ if $m=0$), and define
$$
S(x) =  ( \frac{x}{\omega(x)} \frac{g-1}{g} )^{1/(g-1)} \mbox{ with } \omega(x) = \gamma( (x \frac{g-1}{g} )^{g/(g-1)} ).
$$
Then $s_i = \lceil \frac{1}{d+1}S(\frac{m+i+1}{d+1}) \rceil$ satisfies property (\ref{eqS}) (i.e., (\ref{eqtausample})).
\end{lem}

The lemma simplifies to Lemma \ref{lemsample-simple} if we use $\gamma(z)=1$.

\noindent
{\bf Proof.}
Since $\gamma(z)$ is increasing and $\geq 1$ for $z\geq 0$, also $\omega(x)$ is increasing and $\geq 1$ for $x\geq 0$ (notice that $g>1$). 

We also want to show that $z/\omega(z)$ is increasing for $z\geq 0$. Its derivative is equal to
$$ \frac{1}{\omega(z)}(1-\frac{z \omega'(z) }{\omega(z)})
$$
which is $\geq 0$ if $\omega(z) \geq z \omega'(z) $ (because $\omega(z)$ is positive).
The latter is equivalent (by $\omega$'s definition) to
$$
\gamma( (z \frac{g-1}{g} )^{g/(g-1)} )
\geq z  
\gamma'( (z \frac{g-1}{g} )^{g/(g-1)} )
\frac{g}{g-1} (z \frac{g-1}{g} )^{g/(g-1)-1} \frac{g-1}{g}.
$$
This is implied by $\gamma(y) \geq y \gamma'(y)\frac{g}{g-1}$ for $y= (z \frac{g-1}{g} )^{g/(g-1)}$. From our assumptions on $\gamma(z)$ we infer that this is true. 

Since $x/\omega(x)$ is increasing, also $S(x)$ is increasing for $x\geq 0$. This implies
\begin{eqnarray*}
\sum_{j=0}^i s_j 
&=& \sum_{j=0}^i \lceil \frac{1}{d+1} S(\frac{m+j+1}{d+1}) \rceil
\geq \sum_{j=m}^{i+m} \frac{1}{d+1}S(\frac{j+1}{d+1})  \\
&=& 
\sum_{j=0}^{i+m} \frac{1}{d+1}S(\frac{j+1}{d+1})  - \sum_{j=0}^{m-1} \frac{1}{d+1}S(\frac{j+1}{d+1}) \\
&\geq &\int_{z=0}^{m+i+1} \frac{1}{d+1}S(\frac{z}{d+1}) dz -
\int_{z=0}^{m+1} \frac{1}{d+1}S(\frac{z}{d+1}) dz \\
&=&
\int_{z=0}^{(m+i+1)/(d+1)} S(z) dz -
\int_{z=0}^{(m+1)/(d+1)} S(z) dz.
\end{eqnarray*}

Since $\omega(z)\geq 1$,
$$ \int_{z=0}^{(m+1)/(d+1)} S(z) dz \leq
\int_{z=0}^{(m+1)/(d+1)} (z \frac{g-1}{g})^{1/(g-1)} dz = (\frac{m+1}{d+1} \cdot \frac{g-1}{g})^{g/(g-1)}\leq M_0.
$$
Let $x=(m+i+1)/(d+1)$ 
. Then,
$$ \sum_{j=0}^i s_j  \geq \int_{z=0}^{x} S(z) dz -M_0.$$

Because $\omega(z)$ is increasing for $z\geq 0$ we have
$$\int_{z=0}^{x} S(z) dz
\geq \frac{1}{\omega(x)^{1/(g-1)}} \cdot
\int_{z=0}^{x} (z \frac{g-1}{g})^{1/(g-1)} dz
=
\frac{1}{\omega(x)^{1/(g-1)}} \cdot
(x \frac{g-1}{g})^{g/(g-1)}.
$$ 
Since $x/\gamma(x)$ is increasing (its derivative is equal to $\frac{1}{\gamma(x)}(1-\frac{x \gamma'(x)}{\gamma(x)})$ and is positive by our assumptions on $\gamma(x)$), also $\tau(x)$ is increasing and we infer 
$$ \tau( \sum_{j=0}^i s_j ) \geq
\tau( \int_{z=0}^{x} S(z) dz -M_0)
\geq \tau(\frac{1}{\omega(x)^{1/(g-1)}} \cdot
(x \frac{g-1}{g})^{g/(g-1)} -M_0).$$
By $\tau$'s definition,
the right hand side is equal to
\begin{equation} M_1+
\frac{(\frac{1}{\omega(x)^{1/(g-1)}}(x\frac{g-1}{g})^{g/(g-1)} )^{1/g}}
{\gamma( \frac{1}{\omega(x)^{1/(g-1)}} \cdot
(x \frac{g-1}{g})^{g/(g-1)} )^{1/g} } 
. \label{eqd}
\end{equation}

Since $S(x)$ is increasing, we have
\begin{eqnarray*}
1+\sum_{j=i-d}^i s_j &=& 1+\sum_{j=i-d}^i \lceil \frac{1}{d+1} S(\frac{m+j+1}{d+1})\rceil 
\leq 
1+(d+1) \lceil \frac{1}{d+1} S(\frac{m+i+1}{d+1})\rceil  \\
&\leq& d+2 + S(\frac{m+i+1}{d+1})=d+2+S(x).
\end{eqnarray*}
Since $M_1\geq d+2$, the right hand side is at most (\ref{eqd}) 
if
$$
\frac{(\frac{1}{\omega(x)^{1/(g-1)}}(x\frac{g-1}{g})^{g/(g-1)})^{1/g}}
{\gamma( \frac{1}{\omega(x)^{1/(g-1)}} \cdot
(x \frac{g-1}{g})^{g/(g-1)})^{1/g} }
\geq ( \frac{x}{\omega(x)} \frac{g-1}{g} )^{1/(g-1)} = S(x).
$$
After raising to the power $g$ and reordering terms this is equivalent to

$$ \omega(x)  \geq \gamma( \frac{1}{\omega(x)^{1/(g-1)}} \cdot
(x \frac{g-1}{g})^{g/(g-1)} ).$$
Since $\omega(x)\geq 1$ and $\gamma(z)$ is increasing, this is implied by
$$ \omega(x) \geq 
\gamma( 
(x \frac{g-1}{g})^{g/(g-1)} ).$$

Since this is true by the definition of $\omega(x)$, we obtain inequality  (\ref{eqS}).

\subsection{Diminishing step size sequences} \label{appstep}

We assume the sample size sequence $s_i=\lceil \frac{1}{d+1} S(\frac{m+i+1}{d+1})\rceil$ developed in the previous subsection:

\begin{lem} \label{lemstep}
We assume the sample size sequence $s_i=\lceil \frac{1}{d+1} S(\frac{m+i+1}{d+1})\rceil$ of Lemma \ref{lemsample}.
Let $\{E_t\}$ be a constant or increasing sequence with $E_t\geq 1$. 
For $i\geq 0$, we define
$$ \bar{E}_i = E_{\sum_{j=0}^i s_j}. $$
We define $\bar{E}_{-1}=E_0$ and assume for $i\geq 0$ that
$$ \bar{E}_{i} \leq 2 \bar{E}_{i-1} \mbox{ and } 
s_0-1\leq E_0
.$$

Let $q\geq 0$, and $a_0$ and $a_1$ be constants
such that 
$$ a_1= 
a_0 \cdot \max\{ 3, (1 + \frac{m+2}{m+1})^{1/(g-1)} \}^q.
$$
For $g\geq 2$ and $q\leq 1$ this implies
$a_0\leq a_1 \leq 3\cdot a_0$.

Then 
the diminishing step size sequence $\eta_t = \frac{\alpha_t}{ \mu (t+E_t)^q}$ with parameters $\{\alpha_t \}$  defined by
$$\eta_t = \frac{\alpha_t}{\mu (t+E_t)^q} = \frac{a_0}{\mu ((\sum_{j=0}^{i-1} s_j) +\bar{E}_{i-1})^q} \stackrel{\mbox{{\footnotesize {\sc def}}}}{=} \bar{\eta}_i
$$ 
for $t\in \{(\sum_{j=0}^{i-1} s_j), \ldots, (\sum_{j=0}^{i-1} s_j) +s_i -1 \}$ and $i\geq 0$
 satisfies
$$a_0=\alpha_0\leq \alpha_t \leq a_1.$$
\end{lem}

Notice that Lemma \ref{lemstep-simple} describes the case $g\geq 2$ and $q\leq 1$ in a different wording (by defining a set ${\cal Z}$ of step size sequences).

\noindent
{\bf Proof.} 
For $i\geq 1$, we first show a relation between $s_i$ and $s_{i-1}$ (notice that $\omega$ is increasing):
\begin{eqnarray}
s_i-1 &\leq&
\frac{1}{d+1} S(\frac{m+i+1}{d+1})
=
\frac{1}{d+1} (\frac{1}{\omega(\frac{m+i+1}{d+1})}\frac{m+i+1}{d+1} \frac{g-1}{g} )^{1/(g-1)} \nonumber \\
&\leq &
\frac{1}{d+1} (\frac{1}{\omega(\frac{m+i}{d+1})}\frac{m+i+1}{d+1} \frac{g-1}{g} )^{1/(g-1)} \nonumber \\
&=&
(\frac{m+i+1}{m+i})^{1/(g-1)} \frac{1}{d+1} (\frac{1}{\omega(\frac{m+i}{d+1})}\frac{m+i}{d+1} \frac{g-1}{g} )^{1/(g-1)} \nonumber \\
&=&
(\frac{m+i+1}{m+i})^{1/(g-1)}
\frac{1}{d+1} S(\frac{m+i}{d+1})
\leq 
(\frac{m+2}{m+1})^{1/(g-1)} s_{i-1} \label{sieq}
\end{eqnarray}
For $i\geq 1$ and $t\in \{(\sum_{j=0}^{i-1} s_j), \ldots, (\sum_{j=0}^{i-1} s_j) +s_i -1 \}$, we are now able to derive a bound
\begin{eqnarray*}
\alpha_t &=& a_0 \frac{ (t+E_t)^q }{((\sum_{j=0}^{i-1} s_j) +\bar{E}_{i-1})^{q}}
\leq 
a_0 (\frac{ (\sum_{j=0}^{i} s_j)-1+\bar{E}_{i} }{(\sum_{j=0}^{i-1} s_j) +\bar{E}_{i-1}})^q \\
&\leq &
a_0 (\frac{ (\sum_{j=0}^{i} s_j)-1+2\bar{E}_{i-1} }{(\sum_{j=0}^{i-1} s_j) +\bar{E}_{i-1}})^q
=
a_0 (1 +\frac{ s_i-1+\bar{E}_{i-1}  }{(\sum_{j=0}^{i-1} s_j) +\bar{E}_{i-1}})^q \\
&\leq &
a_0 (1 +\frac{ ((m+2)/(m+1))^{1/(g-1)} s_{i-1}+\bar{E}_{i-1}  }{(\sum_{j=0}^{i-1} s_j) +\bar{E}_{i-1}})^q \\
&\leq&
a_0 (1 +\frac{ ((m+2)/(m+1))^{1/(g-1)} (\sum_{j=0}^{i-1} s_j)+\bar{E}_{i-1}  }{(\sum_{j=0}^{i-1} s_j) +\bar{E}_{i-1}})^q \\
&\leq&
a_0 (1 + (\frac{m+2}{m+1})^{1/(g-1)})^q.
\end{eqnarray*}

Notice that for $g\geq 2$ and $q\leq 1$, this bound is at most
$3\cdot a_0$.

We still need to analyse the case $i=0$. This gives the bound (notice that $s_0-1\leq E_0$ and $\bar{E}_0\leq 2\bar{E}_{-1}=2E_0$)
$$
\alpha_t = a_0 \frac{ (t+E_t)^q }{((\sum_{j=0}^{i-1} s_j) +E_0)^{q}}
=
a_0 \frac{ (t+E_t)^q }{E_0^{q}}
\leq 
a_0 \frac{ (s_0-1+\bar{E}_0)^q }{E_0^{q}}
\leq a_0 \cdot 3^q.
$$

The step size for iteration $t$ is equal to
$$\eta_t = \frac{\alpha_t}{\mu (t+E_t)^q} = \frac{a_0}{\mu ((\sum_{j=0}^{i-1} s_j) +\bar{E}_{i-1})^q},$$
hence, $\alpha_0=a_0$.

\section{Analysis general recursion}

The optimization problem for training many Machine Learning (ML) models using a training set $\{\xi_i\}_{i=1}^M$ of $M$ samples can be formulated as a finite-sum minimization problem as follows
\begin{equation}\label{eq:finite_sum_main}
\min_{w \in \mathbb{R}^d} \left\{ F(w) = \frac{1}{M}
\sum_{i=1}^M f(w; \xi_i) \right\}.
\end{equation}
The objective is to minimize a loss function with respect to model parameters $w$. This problem is known as empirical risk minimization and it covers a wide range of convex and non-convex problems from the ML domain, including, but not limited to, logistic regression, multi-kernel learning, conditional random fields and neural networks.
We are interested in solving the following more general stochastic optimization problem with respect to some distribution $\mathcal{D}$:
\begin{align}
\min_{w \in \mathbb{R}^d} \left\{ F(w) = \mathbb{E}_{\xi \sim \mathcal{D}} [ f(w;\xi) ] \right\},  \label{main_prob_expected_risk}  
\end{align}
where $F$ has a Lipschitz continuous gradient and $f$ has a \emph{finite lower bound} for every $\xi$.

The general form \eqref{main_prob_expected_risk} can be solved by using SGD as described in Algorithm \ref{sgd_algorithm}.
 Thanks to its simplicity in implementation and efficiency in dealing with large scale data sets, stochastic gradient descent, originally introduced in \citep{RM1951}, has become the method of choice for solving not only (\ref{eq:finite_sum_main}) when $m$ is large but also (\ref{main_prob_expected_risk}).

\begin{algorithm}[ht]
  \caption{Stochastic Gradient Descent (SGD) Method}
  \label{sgd_algorithm}
\begin{algorithmic}[1]
  \State {\bfseries Initialize:} $w_0$
  \State {\bfseries Iterate:}
  \For{$t=0,1,2,\dots$}
  \State Choose a step size (i.e., learning rate) $\eta_t>0$. 
  \State Generate a random variable $\xi_t$.
  \State Compute a stochastic gradient $\nabla f(w_{t};\xi_{t}).$
  \State Update the new iterate $w_{t+1} = w_{t} - \eta_t \nabla f(w_{t};\xi_{t})$.
  \EndFor
\end{algorithmic}
\end{algorithm}

\subsection{The Hogwild! algorithm}
\label{app-rec}

To speed up SGD, an asynchronous SGD known as Hogwild! was introduced in~\citep{Hogwild}. 
Here, multiple computing threads work together and update shared memory in asynchronous fashion. The shared memory stores the most recently computed weight as a result of the SGD iterations computed by each of the computing threads.  Writes to and reads from vector positions in shared memory can be inconsistent. As a result a computing thread may start reading positions of the current weight vector from shared memory while these positions are updated by other computing threads out-of-order. Only writes to and reads from shared memory positions are considered atomic. This means that, when a computing thread reads the `current' weight vector from shared memory,  this weight vector is a mix of partial updates to the weight vector from other computing threads that executed previous SGD iterations. 
 
\citep{nguyen2018sgd,nguyen2018new}
introduce a general recursion for $w_t$. The recursion explains which positions in $w_t$ should be updated in order to compute $w_{t+1}$. Since $w_t$ is stored in shared memory and is being updated in a possibly non-consistent way by multiple cores who each perform recursions, the shared memory will contain a vector $w$ whose entries represent a mix of updates. That is, before performing the computation of a recursion, a computing thread will first read  $w$ from shared memory, however, while reading $w$ from shared memory, the entries in $w$ are being updated out of order. The final vector $\hat{w}_t$ read by the computing thread represents an aggregate of a mix of updates in previous iterations.

The general recursion (parts of text extracted from \citep{nguyen2018new}) is defined as follows: For $t\geq 0$,
\begin{equation}
 w_{t+1} = w_t - \eta_t d_{\xi_t}  S^{\xi_t}_{u_t} \nabla f(\hat{w}_t;\xi_t),\label{eqwM}
 \end{equation}
 where
 \begin{itemize}
 \item $\hat{w}_t$ represents the vector used in computing the gradient $\nabla f(\hat{w}_t;\xi_t)$ and whose entries have been read (one by one)  from  an aggregate of a mix of  previous updates that led to $w_{j}$, $j\leq t$, and
 \item the $S^{\xi_t}_{u_t}$ are diagonal 0/1-matrices with the property that there exist real numbers $d_\xi$ satisfying
\begin{equation} d_\xi \mathbb{E}[S^\xi_u | \xi] = D_\xi, \label{eq:SexpM} \end{equation}
where the expectation is taken over $u$ and $D_\xi$ is the diagonal 0/1 matrix whose $1$-entries correspond to the non-zero positions in $\nabla f(w;\xi)$ in the following sense: The $i$-th entry of $D_\xi$'s diagonal is equal to 1 if and only if there exists a $w$ such that the $i$-th position of $\nabla f(w;\xi)$ is non-zero. 
\end{itemize}

The role of matrix $S^{\xi_t}_{u_t}$ is that it filters which positions of gradient $\nabla f(\hat{w}_t;\xi_t)$ play a role in (\ref{eqwM}) and need to be computed. Notice that $D_\xi$ represents the support of $\nabla f(w;\xi)$; by $|D_\xi|$ we denote the number of 1s in $D_\xi$, i.e., $|D_\xi|$ equals the size of the support of $\nabla f(w;\xi)$.

We restrict ourselves to choosing (i.e., fixing a-priori) {\em non-empty} matrices  $S^\xi_u$ that ``partition'' $D_\xi$ in $D$ approximately ``equally sized'' $S^\xi_u$: 
$$ \sum_u S^\xi_u = D_\xi, $$
where each matrix $S^\xi_u$ has either $\lfloor |D_\xi|/D \rfloor$ or $\lceil |D_\xi|/D \rceil$ ones on its diagonal. We uniformly choose one of the matrices $S^{\xi_t}_{u_t}$ in (\ref{eqwM}), hence, $d_\xi$ equals the number of matrices $S^\xi_u$, see (\ref{eq:SexpM}).

In order to explain recursion (\ref{eqwM}) we  consider two special cases. For $D=\bar{\Delta}$, where 
$$ \bar{\Delta} = \max_\xi \{ |D_\xi|\}$$
represents the maximum number of non-zero positions in any gradient computation $f(w;\xi)$, we have that for all $\xi$, there are exactly $|D_\xi|$ diagonal matrices $S^\xi_u$ with a single 1 representing each of the elements in $D_\xi$. Since  $p_\xi(u)= 1/|D_\xi|$ is the uniform distribution, we have $\mathbb{E}[S^\xi_u | \xi] = D_\xi / |D_\xi|$, hence, $d_\xi = |D_\xi|$. This gives the recursion
\begin{equation}
 w_{t+1} = w_t - \eta_t |D_{\xi_t}|  [ \nabla f(\hat{w}_t;\xi_t)]_{u_t},\label{eqwM1}
 \end{equation}
 where $ [ \nabla f(\hat{w}_t;\xi_t)]_{u_t}$ denotes the $u_t$-th position of $\nabla f(\hat{w}_t;\xi_t)$ and where $u_t$ is a uniformly selected position that corresponds to a non-zero entry in  $\nabla f(\hat{w}_t;\xi_t)$.
 
At the other extreme, for $D=1$, we have exactly one matrix $S^\xi_1=D_\xi$ for each $\xi$, and we have $d_\xi=1$. This gives the recursion
\begin{equation}
 w_{t+1} = w_t - \eta_t  \nabla f(\hat{w}_t;\xi_t).\label{eqwM2}
 \end{equation}
Recursion (\ref{eqwM2}) represents Hogwild!. In a single-thread setting where updates are done in a fully consistent way, i.e. $\hat{w}_t=w_t$, yields SGD.

 
 Algorithm \ref{HogWildAlgorithm} gives the pseudo code corresponding to recursion (\ref{eqwM}) with our choice of sets $S^\xi_u$ (for parameter $D$).
 
 \begin{algorithm}
\caption{Hogwild! general recursion}
\label{HogWildAlgorithm}
\begin{algorithmic}[1]

   \State {\bf Input:} $w_{0} \in \mathbb{R}^d$
   \For{$t=0,1,2,\dotsc$ {\bf in parallel}} 
    
  \State read each position of shared memory $w$
  denoted by $\hat w_t$  {\bf (each position read is atomic)}
  \State draw a random sample $\xi_t$ and a random ``filter'' $S^{\xi_t}_{u_t}$
  \For{positions $h$ where $S^{\xi_t}_{u_t}$ has a 1 on its diagonal}
   \State compute $g_h$ as the gradient $\nabla f(\hat{w}_t;\xi_t)$ at position $h$
   \State add $\eta_t d_{\xi_t} g_h$ to the entry at position $h$ of $w$ in shared memory {\bf (each position update is atomic)}
   \EndFor
   \EndFor
\end{algorithmic}
\end{algorithm}

In order to use Algorithm \ref{HogWildAlgorithm} in our asynchronous SGD setting where local compute nodes compute SGD iterations on local data sets, we use the following reinterpretation of shared memory and computing threads.
Compute nodes represent the different computing threads. The centralized server
plays the role of shared memory where locally  computed weight vectors at the compute nodes are `aggregated' as in (\ref{eqwM}). Each ComputeNode$_c$  first reads each of the entries of the weight vector $w$ currently stored at the compute node itself. This weight vector $w$ includes local updates (due to SGD iterations at the compute node itself) as well as updates from other compute nodes since $w$ originally came from the server when transmitting the global model in a broadcast message.
Therefore, ComputeNode$_c$  reads a $\hat{w}_t$ that can be interpreted as a series of successive atomic reads from shared memory  $w$ as described above. Next
the gradient $\nabla f(\hat{w}_t;\xi_t)$ for some sample $\xi_t \sim {\mathcal D}_c$ is computed. Since the order in which compute nodes are executing their SGD iterations is determined by probabilities $\{p_c\}$, this means that from a higher abstraction level $\nabla f(\hat{w}_t;\xi_t)$ for some sample $\xi_t \sim {\mathcal D}$ is computed (that is, with probability $p_c$ sample $\xi_t \sim {\mathcal D}_c$ and ComputeNode$_c$ is the one executing the corresponding SGD iteration). The gradient  multiplied by a step size $\eta_t$ and correction factor $d_{\xi_t}$ is subtracted from the weight vector stored at the server (after it receives the local update from the compute node). This is done by a series of successive atomic writes. Each of the compute nodes (just like the computing threads above) work in parallel continuously updating entries in the weight vector $w$ stored at the server (and stored locally at each compute node). 

In the context of different computing threads atomically reading and writing entries of vector $w$ from shared memory, we define the amount of asynchronous behavior by parameter $\tau$ as in \citep{nguyen2018new}:

\begin{defn} 
We say that weight vector $w$ stored at the server is {\em consistent with delay $\tau$}  with respect to recursion (\ref{eqwM}) if, for all $t$, vector $\hat{w}_t$ includes the aggregate of the updates up to and including those made during the $(t-\tau)$-th iteration (where (\ref{eqwM}) defines the $(t+1)$-st iteration). Each position read from shared memory is atomic and each position update to shared memory is atomic (in that these cannot be interrupted by another update to the same position).
\end{defn}

Even though this (original) definition does not consider $\tau$ as a function of the iteration count $t$, the next subsections do summarize how $\tau$ can depend as a function on $t$.

\subsection{Convergence rate for strongly convex problems}

In this section we 
let $f$ be $L$-smooth, convex, and let the objective function $F(w)=\mathbb{E}_{\xi\sim {\cal D}}[f(w;\xi)]$ be $\mu$-strongly convex with finite $N = 2 \mathbb{E}[ \|\nabla f(w_{*}; \xi)\|^2 ]$ where $w_{*} = \arg \min_w F(w)$.
Notice that we do not assume the bounded gradient assumption which assumes $\mathbb{E}[ \|\nabla f(w; \xi)\|^2 ]$ is bounded for all $w\in \mathbb{R}^d $ (not only $w=w_{*}$ as in Assumption \ref{ass_finitesigma}) and is in conflict with assuming strong convexity as explained in \citep{nguyen2018sgd,nguyen2018new}.

\subsubsection{Constant step sizes} \label{appconst}
Algorithm~\ref{HogWildAlgorithm} for $D=1$ corresponds to Hogwild! with recursion (\ref{eqwM2}). For finite-sum problems,
\citep{Leblond2018}
 proves for 
  constant step sizes $\eta_t=\eta=\frac{a}{L}$ with delay 
\begin{equation}
\tau\leq \frac{1}{\eta\mu}
\label{tau2}
\end{equation}
  and parameter 
  $a\leq (5(1+2\tau \sqrt{\Delta})\sqrt{1+\frac{\mu}{2L}\min\{\frac{1}{\sqrt{\Delta}},\tau\}})^{-1}$ as a function of $\tau$, where $\Delta$ measures  sparsity according to Definition 7 in~\citep{Leblond2018}, 
that the convergence rate $\mathbb{E}[\|\hat{w}_t - w_*\|^2]$ is at most
$$
    \mathbb{E}[\|\hat{w}_t - w_*\|^2]\leq 2 (1-\rho)^t \|w_0-w_*\|^2 + b,
    $$
where $\rho=\frac{a L}{\mu}$ and $b=(\frac{4\eta(C_1+\tau C_2)}{\mu} + 2\eta^2C_1\tau )N$ for 
$C_1= 1 + \sqrt{\Delta}\tau$ and 
$C_2= \sqrt{\Delta} + \eta \mu C_1$.

With a fixed learning rate $\eta_t = \eta$, SGD and Hogwild! may provide fast initial improvement, after which it oscillates within a region containing a solution \citep{BottouCN18,CheeT18}. The upper bound on the convergence rate shows that convergence is to within some range of the optimal value; we have $\mathbb{E}[\|\hat{w}_t - w_*\|^2]= \mathcal{O}(\eta)$.

 Hence, SGD and Hogwild! can fail to converge to a solution. It is known that the behavior of SGD is strongly dependent on the chosen learning rate and on the variance of the stochastic gradients. To overcome this issue, there are two main lines of research that have been proposed in literature: variance reduction methods 
 and diminishing learning rate schemes. 
 These algorithms guarantee to converge to the optimal value.   
 
 Upper bound (\ref{tau2}) allows us to set the maximal sample size $s=s_i$  in terms the amount of asynchronous behavior allowed by $d$ in (\ref{eqS}). This allows an informed decision on how to reduce the number of broadcast messages as much as possible. We need $\tau=(d+1) s \leq \frac{1}{\eta \mu}$, i.e., $s\leq \frac{1}{\eta \mu (d+1)}= \frac{a}{L \mu  (d+1)}$ which is typically large.

\subsubsection{Diminishing step sizes} \label{app-sc-dim}

We slightly reformalize\footnote{The original lemma requires $2L\alpha/\mu \leq \tau(t)\leq \sqrt{\frac{t}{\ln t}  \cdot \left( 1 - \frac{1}{\ln t} \right)}$ which cannot be realized for small $t$. Nevertheless, if $\tau(t)\leq \sqrt{\frac{t}{\ln t}  \cdot \left( 1 - \frac{1}{\ln t} \right)}$ for $t\geq T_1$ where $T_1$ is a constant, then the derivation after (47) in \citep{nguyen2018sgd} still holds true if we consider $\sum_{i=T_1}^t a_i \tau(i)^2$ and $\sum_{i=T_1}^t a_i \tau(i)$, respectively. We can replace the sums $\sum_{i=1}^{T_1} a_i \tau(i)^2$ and $\sum_{i=1}^{T_1} a_i \tau(i)$ by $O(1)$ terms and as a result the derivations that lead to Lemma 11 still follow.} Lemma 11 from \citep{nguyen2018sgd}:

\begin{lem} \label{lemsc} \citep{nguyen2018sgd,nguyen2018new}
 Let $\tau(t)$ be a delay function satisfying  $2L\alpha/\mu \leq \tau(t)\leq t$ with the additional property that there exists a constant $T_1$ such that for large enough $t\geq T_1$, $\tau(t)\leq  \sqrt{\frac{t}{\ln t}  \cdot \left( 1 - \frac{1}{\ln t} \right)}$.
Let 
$$\{\eta_t = \frac{\alpha_t}{\mu (t+2\tau(t) )} \}$$ 
be a step size sequence with $12\leq \alpha_t\leq \alpha$. 
Then the expected convergence rates satisfy
\begin{eqnarray*}
 \mathbb{E}[\|\hat{w}_{t+1} - w_* \|^2] &\leq &  \frac{4\alpha^2 D N}{\mu^2}\frac{1}{t} + O(\frac{1}{t\ln t})  \mbox{ and } \\
 \mathbb{E}[\|w_{t+1} - w_* \|^2] &\leq&  \frac{4\alpha^2 D N}{\mu^2}\frac{1}{t} + O(\frac{1}{t\ln t}).
 \end{eqnarray*}
\end{lem}

We apply Lemma \ref{lemsc} for $D=1$ which corresponds to Hogwild! as applied in this paper.

The conditions of Lemma \ref{lemsample} are satisfied for $g=2$ and $\gamma(z) = 4\ln z$. Let 
$$ M_1=\max \left\{d+2, \frac{2L\alpha}{\mu}, 
\frac{1}{2}\left\lceil \frac{m+1}{16(d+1)^2} \frac{1}{ \ln(\frac{m+1}{2(d+1)})} \right\rceil
\right\}$$ 
and 
$$M_0=\left((m+1) \frac{g-1}{g}\right)^{g/(g-1)} = \frac{(m+1)^2}{4}.
$$
This defines
$$\tau(t) = M_1 + \left(\frac{t+M_0}{\gamma(t+M_0)}\right)^{1/g}=M_1+ \sqrt{\frac{t+M_0}{ 4 \ln(t+M_0)}}.$$
For $t$ large enough such that both $t\geq \max\{M_0, e^3\}$ and $(1-\sqrt{3/4})\cdot\sqrt{(t+ M_0)/ \ln(t+ M_0) }\geq M_1$, $\tau(t)\leq \sqrt{(t/\ln t)\cdot (1-1/\ln t)}$ for the following reason:
The square root  
$$\sqrt{\frac{t+M_0}{ 4 \ln(t+M_0)}}\leq \sqrt{\frac{3}{4} \frac{t}{\ln t}  \cdot \left( 1 - \frac{1}{\ln t} \right)}$$
because $1-1/\ln t \geq 2/3$ and $(t+M_0)/\ln(t+M_0) \leq (t+M_0)/\ln t \leq 2t/\ln t$.
Adding the bound on $M_1$ completes the argument. Hence, we can apply Lemma \ref{lemsc}. We use Lemma \ref{lemsample} to obtain a concrete sample size sequence $\{s_i\}$.
This leads to an increasing sample size sequence defined by
$$ s_i = \lceil \frac{m+i+1}{16(d+1)^2} \frac{1}{ \ln(\frac{m+i+1}{2(d+1)})} \rceil =
O(\frac{i}{\ln i}).
$$
(For example, $s_0=\lceil 31.989/(d+1) \rceil$  corresponds to $m+1=2(d+1)\cdot 1937$. For $d=1$, this gives $s_0=16$.)

By setting $E_t=2\tau(t)$ in the step size sequence based on $\{\eta_t\}$ of Lemma \ref{lemsc}, we can apply our recipe of Lemma \ref{lemstep} for computing the round step size sequence $\{\bar{\eta}_i\}$. 
Notice that we need to choose $a_0=12$ and $\alpha=a_1\leq 3\cdot a_0=36$ in Lemma  \ref{lemstep}; we use $\alpha=36$ in the formulas below. Also notice that $E_0=2\tau(0)\geq s_0$ by the definitions of $s_0$, $M_1$ and $\tau(t)$.

The inequality $\bar{E}_{i+1} \leq 2 \bar{E}_i$ follows from (remember $E_t=2\tau(t)$)
$$ \sqrt{\frac{\sum_{j=0}^{i} s_j +M_0}{4 \ln ( \sum_{j=0}^{i} s_j +M_0) }}
\leq 2 \sqrt{\frac{\sum_{j=0}^{i-1} s_j +M_0}{4 \ln ( \sum_{j=0}^{i-1} s_j +M_0) }}.
$$
Notice that (\ref{sieq}) for $g=2$  implies $s_{i} -1 \leq 2s_{i-1}$. Since $s_{i-1}\geq 1$ we have $s_i \leq 3 s_{i-1}$. We derive
\begin{eqnarray*}
\sqrt{\frac{\sum_{j=0}^{i} s_j +M_0}{4 \ln ( \sum_{j=0}^{i} s_j +M_0) }}
&\leq & 
\sqrt{\frac{\sum_{j=0}^{i} s_j +M_0}{4 \ln ( \sum_{j=0}^{i-1} s_j +M_0) }}
\leq
\sqrt{\frac{\sum_{j=0}^{i-1} s_j +3 s_{i-1}+ M_0}{4 \ln ( \sum_{j=0}^{i-1} s_j +M_0) }} \\
&\leq &
\sqrt{\frac{4(\sum_{j=0}^{i-1} s_j + M_0)}{4 \ln ( \sum_{j=0}^{i-1} s_j +M_0) }}
=
2 \sqrt{\frac{\sum_{j=0}^{i-1} s_j +M_0}{4 \ln ( \sum_{j=0}^{i-1} s_j +M_0) }}.
\end{eqnarray*}
It remains to show that $\bar{E}_1\leq 2 E_0=2 \bar{E}_{-1}$. This follows from a similar derivation as the one above if we can show that $s_0\leq 3 M_0$. The latter is indeed true because $\frac{1}{16(d+1)^2}\leq \frac{1}{16} \leq \frac{3}{4} \leq 3 \frac{m+1}{4}$.

Now we are ready to apply Lemma \ref{lemstep} from which we obtain  diminishing round step size sequence defined by
$$
\bar{\eta}_i = \frac{12}{\mu} \cdot \frac{1}
{\sum_{j=0}^{i-1} s_j + 
2M_1 
+
\sqrt{\frac{(m+1)^2/4+\sum_{j=0}^{i-1} s_j}{\ln((m+1)^2/4+ \sum_{j=0}^{i-1} s_j)}}}
%
= O(\frac{\ln i}{i^2}).
$$

Application of Lemma \ref{lemsc} {\em proves Theorem \ref{thmsc2}} (in the main body).

\subsubsection{Tightness} \label{sec:tight}

For first order stochastic algorithms the best  attainable convergence rate for $t$ is at least $\frac{1}{2} \frac{N}{\mu^2} \frac{1}{t} (1- O((\ln t)/t))$ as shown  in \citep{nguyen2019tight}. This means that for increasing $t$ the upper bound on the expected convergence rate of  Lemma \ref{lemsc} converges to factor $\leq 2 \cdot 4 \cdot 36^2 = 10368$ times the best attainable convergence rate. 

The factor 10368 can be improved to a smaller value: The component 36 comes from $3a_0$ where the factor 3 follows from coarse upper bounding in the proof of Lemma \ref{lemstep}.

\vspace{3mm}

\noindent
{\bf Discussion on how fast the upper bound gets close to within a constant factor of the best attainable convergence rate:}
The $O(1/(t \ln t))$ term in the upper bound and $O((\ln t)/t)$ term in the lower bound (given by the best attainable convergence rate) converges to 0, but how fast? Some indication follows from analysis provided in \citep{nguyen2018sgd} where is stated that 
 if $\tau(t)$ is set to meet the upper bound $\sqrt{(t/\ln t)\cdot (1-1/\ln t)}$, then for large enough 
$$t\geq T_0 =  \exp[ 2\sqrt{\Delta}(1+\frac{(L+\mu)\alpha}{\mu})]$$
the constants of all the asymptotic higher order terms in $O(1/(t \ln t))$ in the upper bound on the convergence rate that contain $\tau(t)$ are such that the concrete values of all these terms are at most the leading explicit term of the upper bound on the convergence rate.\footnote{For completeness, the leading term is also independent of $\|w_0-w_*\|^2$ -- it turns out that for $t \geq T_1 = \frac{\mu^2}{\alpha^2 N D}\|w_0-w_*\|^2$ the higher order term that contains $\|w_0-w_*\|^2$ is at most the leading term.}
So, for $t\geq T_0$,
we have that the upper bound on the convergence rate starts to look like 2 (or a small factor) times the leading explicit term without the asymptotic $O(1/(t \ln t))$ term. This gives an indication when the upper bound starts to get close to $2\cdot 4\cdot 36^2$ times the best attainable convergence rate.

Our recipe in Lemma \ref{lemsample-simple} for computing a sample size sequence $\{s_i\}$ can be applied for a smaller delay function, e.g., $\tau(t)=t^{1/3}$. For such a new sample size sequence we can repeat the same type of calculations as presented in this subsection and show how the diminishing round step size sequence  computed according to our recipe in Lemma \ref{lemstep}  can still achieve an $O(1/t)$ upper bound on the convergence rate which itself converges within a constant factor of the best attainable convergence rate. This convergence sets in for $t\geq T_0$ where $T_0$ corresponds to $\tau(t)=t^{1/3}$. The new $T_0$ is expected to be significantly smaller. So, the $O(1/t)$ convergence rate sooner gets close to within a constant factor of the best attainable convergence rate  if we use a less increasing sample size sequence. In theory this means (1) less rounds because 
the convergence rate diminishes faster,
hence, a smaller number $K$ of total gradient computations is needed. And (2), this means more rounds because the sample sizes are chosen smaller while still a total number of $K$ gradient computations are needed for the desired convergence. It is an open problem to find the right theoretical balance. In experiments, however, we see that even with linear increasing sample size sequences we attain (fast) practical convergence. So, the offered theoretical analysis in this subsection should be used as an understanding that using a  (linearly) increasing sample size sequence is a promising design trick; 
experiments show  practical convergence not only for strongly convex problems but also for plain and non-convex problems. 

\subsection{Plain convex and non-convex problems} \label{app-nonconvex}

For an objective function $F$ (as defined in (\ref{eqObj})), we are generally interested in the expected convergence rate
\begin{equation*}
Y^{(F)}_t =  \mathbb{E}[ F(w_t)-F_{*}],
\end{equation*}
 where $F_{*} = F(w_{*})$ for a global minimum $w_{*}$ (and the expectation is over the randomness used in the recursive computation of $w_t$ by the probabilistic optimization algorithm of our choice). 
This implicitly assumes that there exists a global minimum $w_{*}$, i.e., $\mathcal{W}^*=\{ w_{*} \in \mathbb{R}^d \ : \ \forall_{w\in \mathbb{R}^d} \ F(w_{*})\leq F(w)\} $ defined as the set  of all $w_{*}$ that minimize $F(\cdot)$ is non-empty. 
Notice that $\mathcal{W}^*$ can have multiple global minima even for convex problems. 

For convex problems a more suitable definition for $Y_t$ is the averaged expected convergence rate defined as 
\begin{equation*}
Y^{(A)}_t = \frac{1}{t+1} \sum_{i=t+1}^{2t} \mathbb{E}[ F(w_t)-F_{*} ]
\end{equation*}
and  for strongly convex objective functions we may use 
\begin{equation*}
Y^{(w)}_t = \mathbb{E}[ \mbox{inf} \{ \Vert w_t - w_{*}\Vert ^2 \ : \ w_{*} \in \mathcal{W}^* \} ].
\end{equation*}
The $\omega$-convex objective functions define a range of functions between plain convex and strongly convex for which both convergence rate definitions can be computed for diminishing step sizes~\citep{van2019characterization}. From \citep{van2019characterization} we have that an objective function with 'curvature' $h\in [0,1]$ (where $h=0$ represents plain convex and $h=1$ represents strongly convex) achieves convergence rates $Y_t^{(w)}=O(t^{-h/(2-h)})$ and $Y_t^{(A)}=O(t^{-1/(2-h)})$ for diminishing step sizes $\eta_t = O(t^{-1/(2-h)})$. For this reason, when we study plain convex problems, we use  diminishing step size sequence $O(t^{-1/2})$ and we experiment with different increasing sample size sequences to determine into what extent asynchronous SGD or Hogwild! is robust against delays. Since strongly convex problems have best convergence and therefore best robustness against delays, we expect a suitable increasing sample size sequence $O(i^p)$ for some $0\leq p\leq 1$.

For non-convex (e.g., DNN) problems we generally use the averaged expected squared norm of the objective function gradients:
 \begin{equation*}
 Y^{(\nabla)}_t=\frac{1}{t+1} \sum_{j=0}^t  \mathbb{E}[\Vert \nabla F(w_j) \Vert ^2].
 \end{equation*}
This type of convergence rate analyses convergence to a stationary point, which is a candidate for any (good or bad)  local minimum and does not exclude saddle-points. 
There may not even exist a global minimum (i.e., ${\cal W}^*$ is empty) in that some entries in the weight vector $w_t$ may tend to $\pm \infty$ -- nevertheless, there may still exist a value $F_{*} = \mbox{sup} \{ F_{low} \ :  F_{low}\leq F(w), \  \forall{w\in \mathbb{R}^d}\}$ which can be thought of as the value of the global minimum if we include limits to infinite points. In practice a diminishing step size sequence of $O(t^{-1/2})$ (as in the plain convex case)  gives good results and this is what is used in our experiments.

\section{Experiments}
\label{app:experiment}

In this section, we provide  experiments to support our theoretical findings, i.e., the convergence of our proposed asynchronous SGD with strongly convex, plain convex and non-convex objective functions.

We introduce the settings and parameters used in the experiments in Section~\ref{subsec:hyperparameter}. Section~\ref{subsec:exp_asynFL} provides  detailed experiments for asynchronous SGD with different types of objective functions (i.e., strongly convex, plain convex and non-convex objective functions), different types of step size schemes (i.e., constant and diminishing step size schemes), different types of sample size sequences (i.e., constant and increasing sample sizes) and different types of data sets at the clients' sides (i.e., unbiased and biased).

Our experiments are mainly conducted  on LIBSVM\footnote{https://www.csie.ntu.edu.tw/~cjlin/libsvmtools/datasets/binary.html} and MNIST data sets.

\subsection{Experiment settings}
\label{subsec:hyperparameter}

\vspace{0.2cm}
\noindent
\textbf{Simulation environment.} For simulating the asynchronous SGD, we use multiple threads where each thread represents one compute node joining the training process. The experiments are conducted on Linux-64bit OS, with $16$ cpu processors and 32Gb RAM. 


\vspace{0.2cm}
\noindent
\textbf{Experimental setup.} Equation~(\ref{eq_logstic_reg}) defines the plain convex logistic regression problem. The weight vector $\bar{w}$ 
and bias value $b$ 
of the logistic function can be learned by minimizing the log-likelihood function $J$:
\begin{equation} \label{eq_logstic_reg}
    J = - \sum_{i}^M [ y_{i} \cdot \log (\sigma_i) + (1 - y_{i}) \cdot \log (1 - \sigma_i) ], \text{ (plain convex)}
\end{equation}
where $M$ is the number of training samples $(x_i,y_i)$ with $y_i\in\{0,1\}$, $\sigma_i=\sigma(\bar{w},b,x_i)$ and
\begin{equation} \nonumber 
    \sigma(\bar{w}, b, x) = \frac{1}{1 + e^{-(\bar{w}^{\mathrm{T}}x + b)}}
\end{equation}
is the sigmoid function with as parameters the weight vector $\bar{w}$ and bias value $b$. The goal is to learn a vector $w^*$ which represents a pair $w=(\bar{w}, b)$  that minimizes $J$. 

Function $J$ can be changed into a strongly convex problem by adding a regularization parameter $\lambda>0$:
\begin{equation} \nonumber 
    \hat{J} = - \sum_{i}^M [ y_{i} \cdot \log (\sigma_i) + (1 - y_{i}) \cdot \log (1 - \sigma_i) ] + \frac{\lambda}{2}\norm{w}^{2}, \text{ (strongly convex),}
\end{equation}
where $w=(\bar{w},b)$ is  vector $\bar{w}$ concatenated with  bias $b$.
For simulating non-convex problems, 
we choose a simple neural network (LeNet) \citep{lecun1998gradient} for image classification.

The parameters used for our asynchronous SGD algorithm with strongly convex, plain convex and non-convex objective functions are described in Table~\ref{tbl:tbl_async_fl_parameter}.

\begin{table}[H]
\caption{Default asynchronous SGD training parameters}
\label{tbl:tbl_async_fl_parameter}
\vskip 0.1in
\begin{center}
\begin{small}
\scalebox{0.99}{
\begin{threeparttable}

\begin{tabular}{|l|c|c|c|c|c|}
\hline
                 & Sample size sequence & \# of compute nodes & Diminishing step size $\eta_t$ & Reg. par. $\lambda$         \\ \hline
Strongly convex  & $s_i\tnote{\dag} = a \cdot i^c + b$             & 5                 & $\frac{\eta_0}{1 + \beta {t}\tnote{\ddag}}$                & $\frac{1}{M}$                  \\ \hline
Plain convex  & $s_i = a \cdot i^c + b$             & 5                 & $\frac{\eta_0}{1 + \beta {t}}$ or $\frac{\eta_0}{1 + \beta \sqrt{t}}$ &  $N/A$          \\ \hline
Non-convex    & $s_i = a \cdot i^c + b$             & 5                & $\frac{\eta_0}{1 + \beta \sqrt{t}}$ &  $N/A$ \\ \hline
\end{tabular}
 \begin{tablenotes}
       \item {\footnotesize $\dag$ This is the total sample size for the $i$-th communication round,  i.e., $s_i= \sum_{c=1}^n s_{i,c}$ where $s_{i,c}$ is the sample size of client $c\in \{1,\ldots, n\}$. By default $s_{i,c}=s_i/n$}. 
       \item {\footnotesize $\ddag$ The $i$-th round step size $\bar{\eta}_i$ is computed by substituting  $t=\sum_{j=0}^{i-1} s_j$ into the diminishing step size formula}.
   \end{tablenotes}
  \end{threeparttable}

}
\end{small}
\end{center}
\vskip -0.1in
\end{table}

For plain convex problems, we may use the diminishing step size schemes $\frac{\eta_0}{1 + \beta \cdot t}$ or $\frac{\eta_0}{1 + \beta \cdot \sqrt{t}}$, although our experiments in this paper focus on using  $\frac{\eta_0}{1 + \beta \cdot \sqrt{t}}$.

In our experiments we use  parameter $\beta=0.001$ for strongly convex problems and $\beta=0.01$ for plain and non-convex problems.
Parameter $\eta_0$ is the initial step size which we compute by performing a systematic grid search for $\beta$ (i.e., we select the $\eta_0$ giving 'best' convergence).



When we talk about accuracy (in Table \ref{tbl_async_fl_strongly_convex_constant_sample_size} and onward), we mean test accuracy defined as the the fraction of samples from a test data set that get accurately labeled by the classifier (as a result of training on a training data set by minimizing a corresponding objective function).
Test error in Fig \ref{fig:async_fl_mnist_neural_net}, \ref{fig:asyn_fl_biased_mnist}, \ref{fig:async_fl_biased_non_convex_mnist} means one minus the test accuracy.

Rather than test accuracy, we may measure convergence by plotting $F(w_t) - F(w^{*})$, where $F$ is the objective function (corresponding to $J$ and $\hat{J}$ for strongly and plain convex problems, and corresponding to LeNet for the non-convex problem of image classification).  The value $F(w_t)-F(w^*)$ reflects how close the $t$-th iteration gets to the minimal objective function value. Here, we estimate the actual minimum $w^*$ by using a single SGD with diminishing step sizes
for a very large number of iterations .


\subsection{Asynchronous SGD}
\label{subsec:exp_asynFL}

We consider our asynchronous SGD with strongly convex, plain convex and non-convex objective functions for different settings, i.e., different step size schemes (constant and diminishing step size sequences), different sample size schemes (constant and increasing samples size sequences), and different type of the data sets (biased and unbiased data sets across compute nodes).

\subsubsection{Asynchronous SGD with constant step size sequence and constant sample size sequence}
\label{subsec:constant_stepsize_samplesize_FL_simu}

The purpose of this experiment is to find 
the best constant sample size sequence we can use. 
We will use this in Section 
\ref{subsec:stepsize_FL_simu}
to compare constant step size and constant sample size sequences with diminishing step size sequences that use increasing sample sizes. 

For simplicity, we set the total number of iterations at $K=20,000$  for $n=5$ compute nodes, and a constant step size $\eta=0.0025$.

\begin{table}[ht!]
\caption{The accuracy of asynchronous SGD with constant sample sizes}
\label{tbl_async_fl_strongly_convex_constant_sample_size}
\vskip 0.1in

\begin{center}
\begin{footnotesize}
\scalebox{0.99}{
\begin{tabular}{|c|l|cccccc|}
\hline
& Sample size          & 50      & 100     & 200     & 500     & 700     & 1000    \\
& \# of communication rounds & 80      & 40      & 20      & 8       & 6       & 4       \\
\hline \hline
{a9a}  & strongly convex             & 0.8418 & \textbf{0.8443} & \textbf{0.8386} & 0.8333 & 0.8298 & 0.7271 \\

{} & plain convex             & 0.8409 & \textbf{0.8415} & \textbf{0.8417} & 0.8299 & 0.8346 & 0.7276 \\

\hline \hline

{covtype.binary}  & strongly convex            & 0.8429 & \textbf{0.8402} & \textbf{0.8408} & 0.8360 & 0.8278 & 0.7827 \\

{}  & plain convex               & 0.8421 & \textbf{0.8438} & \textbf{0.8404} & 0.8381 & 0.8379 & 0.7287 \\

\hline \hline

{mnist}  & non-convex               & 0.9450 & \textbf{0.9470} & \textbf{0.9180} & 0.8740 & 0.8700 & 0.7080 \\

\hline
\end{tabular}
}
\end{footnotesize}
\end{center}
\vskip -0.1in
\end{table}

The results are in Table~$\ref{tbl_async_fl_strongly_convex_constant_sample_size}$. We see that with  constant sample size $s_i=100$ or $s_i=200$, we get the best accuracy for strongly convex, plain convex and non-convex cases, when compared to the other constant sample sizes. We notice that we can choose the small constant sample size $s_i=50$ as well, because this  also achieves good accuracy, however, this constant sample size requires the algorithm to run for (much) more communication rounds. In conclusion, we need to choose a suitable constant sample size, for example $s_i=100$ or $s_i=200$ to get a good accuracy and a decent number of communication rounds.

\subsubsection{Asynchronous SGD with diminishing step size sequence and increasing sample size sequence}
\label{subsec:samplesize_FL_simu}

To study the behaviour of asynchronous SGD with different increasing sample size sequences, we conduct the following experiments for both $O(i)$ and $O(\frac{i}{\ln i})$ sequences. The purpose of this experiment is to show that these two sampling methods provide good accuracy.

\begin{figure}[ht!]
\begin{center}
\includegraphics[width=0.5\textwidth]{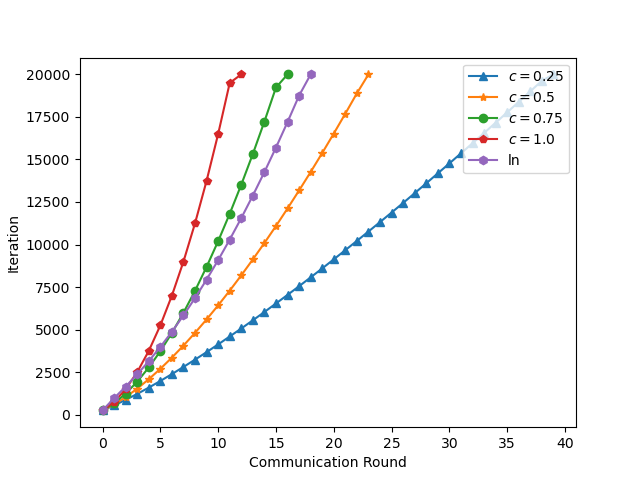}
\end{center}
\caption{Number of communication rounds by sampling methods.}
\label{fig:async_fl_sampling_method}
\end{figure}

\begin{figure}[ht!]
  \centering
  \subfloat[Strongly convex.]{\includegraphics[width=0.5\textwidth]{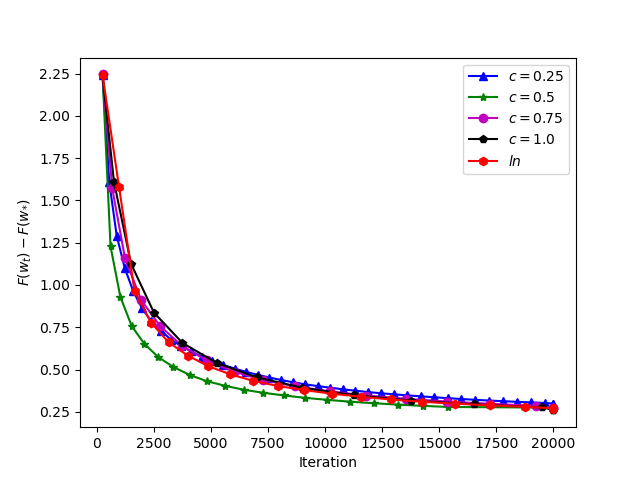}\label{fig:async_fl_a9a_sampling_method_1}}
  \hfill
  \subfloat[Plain convex.]{\includegraphics[width=0.5\textwidth]{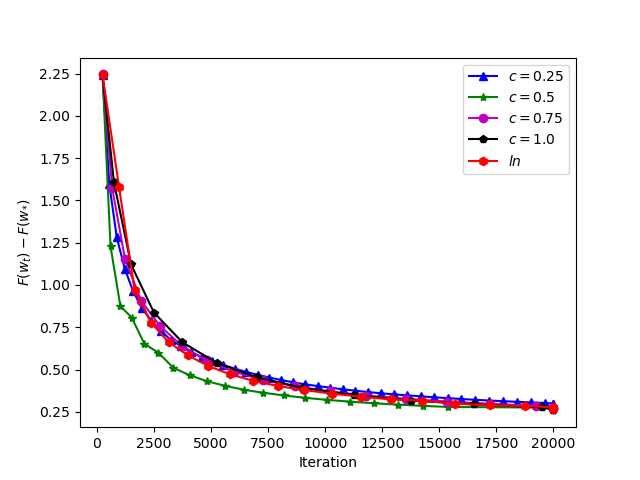}\label{fig:async_fl_plain_convex_a9a_sampling_method_2}}
  \caption{Effect of sampling methods (a9a dataset).}
  \label{fig:async_fl_a9a_sampling_method}
\end{figure}

\begin{figure}[ht!]
  \centering
  \subfloat[Strongly convex.]{\includegraphics[width=0.5\textwidth]{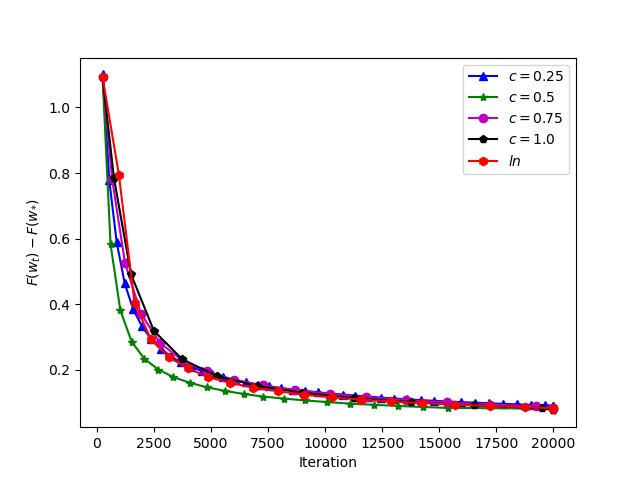}\label{fig:async_fl_covtype_binary_sampling_method_1}}
  \hfill
  \subfloat[Plain convex.]{\includegraphics[width=0.5\textwidth]{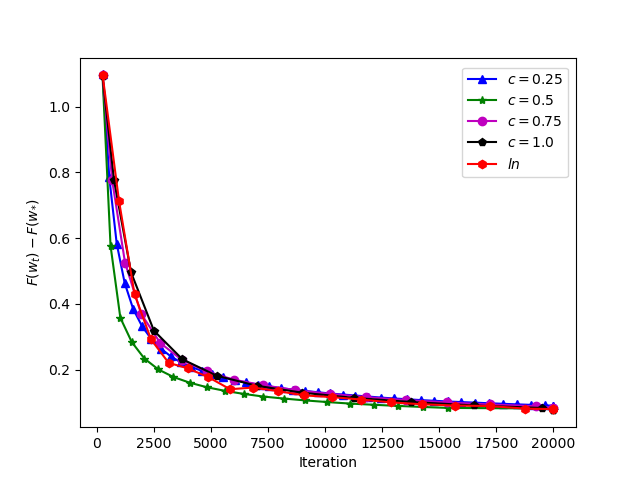}\label{fig:async_fl_plain_convex_covtype_binary_sampling_method_2}}
  \caption{Effect of sampling methods (covtype-binary dataset).}
  \label{fig:async_fl_covtype_binary_sampling_method}
\end{figure}

\textbf{Sampling method:} In this paper, we choose two ways to increase the sample sizes from one communication round to the next. Let  $s_i$ be the number of iterations that the collection of all compute nodes runs in round $i$:
\begin{enumerate}
    \item{${O}(i)$ method: $s_i = a \cdot i^c + b$ where $c \in [0, 1]$, $a,b \geq 0$. 
    }
    
     \item{${O}(\frac{i}{\ln i})$ method: $s_i = a \cdot \frac{i}{\ln(i)} + b$ where $a, b \geq 0$.}
\end{enumerate}

For simplicity, we set $b=0, a=50$, and the total number of iterations at $20,000$ for $5$ compute nodes in Figure~$\ref{fig:async_fl_sampling_method}$. We choose the diminishing step size sequence $\frac{\eta_0}{1 + \beta \cdot t}$ for the strongly convex case and $\frac{\eta_0}{1 + \beta \cdot \sqrt{t}}$ for the plain convex case, with initial step size $\eta_{0} = 0.01$. From Figure~$\ref{fig:async_fl_a9a_sampling_method}$ and Figure~$\ref{fig:async_fl_covtype_binary_sampling_method}$ we infer that $O(i)$ with $c=1.0$ and $O(\frac{i}{\ln{i}})$ sample size sequences provide  fewer communication rounds while maintaining a good accuracy, when compared to other settings. (For completeness, we verified that other increasing sample size sequences, such as exponential increase and cubic increase, are not  good choices for our asynchronous SGD setting.)

\subsubsection{Comparison  asynchronous SGD with (constant step size, constant sample size) and (diminishing step size, increasing sample size).}
\label{subsec:stepsize_FL_simu}

In order to understand the behavior of the asynchronous SGD for different types of step sizes, we conduct the following experiments on strongly convex, plain convex and non-convex problems.


\textbf{Diminishing step size scheme:} We use the following two diminishing step size schemes:
\begin{enumerate}
    \item{Diminishing step size scheme over iterations (diminishing$_1$): Each compute node $c\in \{1,\ldots, n\}$ uses $\eta_t$ for $t = \sum_{c=1}^n (\sum_{j=0}^{i-1} s_{j,c} + h)= \sum_{j=0}^{i-1} s_{j} + n\cdot h$ where $i \geq 0$ denotes the current round, and $h \in \{0,\ldots,  s_{i,c}-1\}$ is the current iteration which the compute node executes. In our experiments all compute nodes use the same sample sizes $s_{j,c}=s_j/n$.} 
    
     \item{Diminishing step size scheme over rounds (diminishing$_2$): Each client $c$ uses a round step size $\bar{\eta}_i$ for all iterations in round $i$. The round step size $\bar{\eta}_i$ is equal to $\eta_t$ for  $t = \sum_{j=0}^{i-1} s_{j}$.
     }
     
\end{enumerate}

\begin{figure}[ht!]
  \centering
  \subfloat[Strongly convex.]{\includegraphics[width=0.5\textwidth]{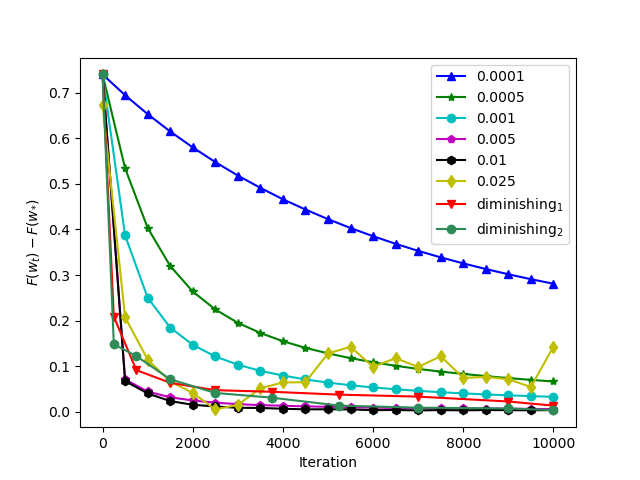}\label{fig:async_fl_strongly_convex_phishing_convergence}}
  \hfill
  \subfloat[Plain convex.]{\includegraphics[width=0.5\textwidth]{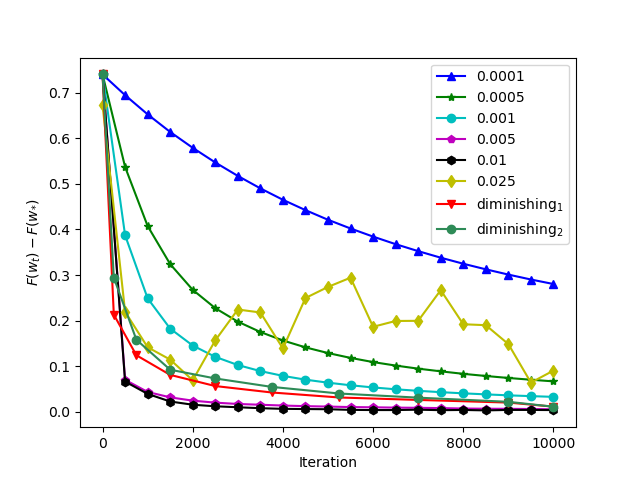}\label{fig:async_fl_plain_convex_phishing_convergence}}
  \caption{Convergence rate with different step sizes (phishing dataset)}
  \label{fig:async_fl_phishing_convergence}
\end{figure}

\begin{figure}[ht!]
  \centering
  \subfloat[Strongly convex.]{\includegraphics[width=0.5\textwidth]{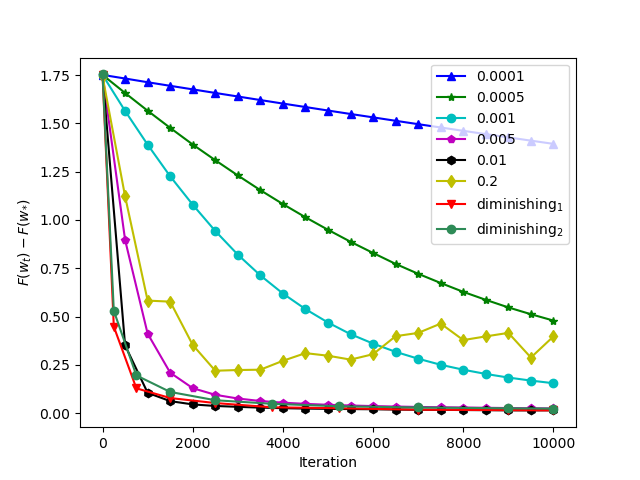}\label{fig:async_fl_strongly_convex_ijcnn1_convergence}}
  \hfill
  \subfloat[Plain convex.]{\includegraphics[width=0.5\textwidth]{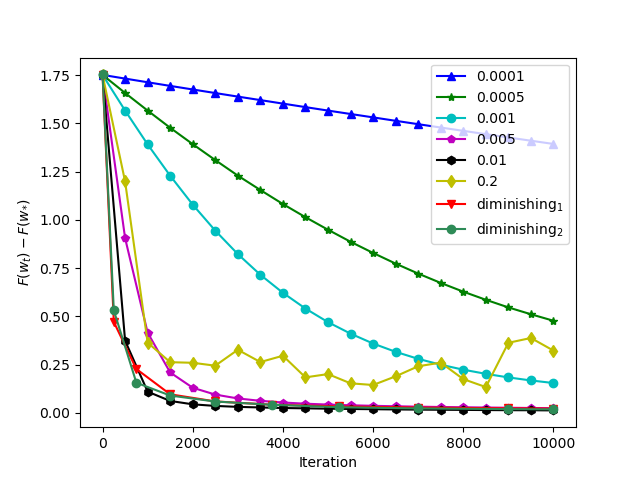}\label{fig:async_fl_plain_convex_ijcnn1_convergence}}
  \caption{Convergence rate with different step sizes (ijcnn1 dataset)}
  \label{fig:async_fl_ijcnn1_convergence}
\end{figure}

\begin{figure}[ht!]
\begin{center}
\includegraphics[width=0.5\textwidth]{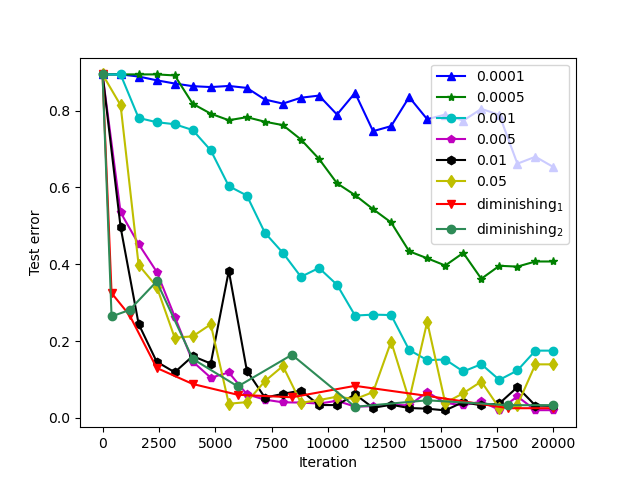}
\end{center}
\caption{Asynchronous SGD in non-convex (MNIST dataset)}
\label{fig:async_fl_mnist_neural_net}
\end{figure}

\begin{figure}[ht!]
  \centering
  \subfloat[Strongly convex (real-sim dataset).]{\includegraphics[width=0.5\textwidth]{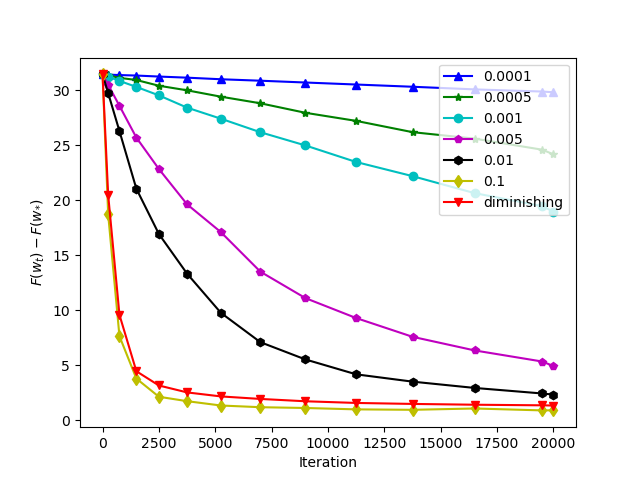}\label{fig:async_fl_stepsize_method_decay_vs_constant_real_sim}}
  \hfill
  \subfloat[Strongly convex (w8a dataset).]{\includegraphics[width=0.5\textwidth]{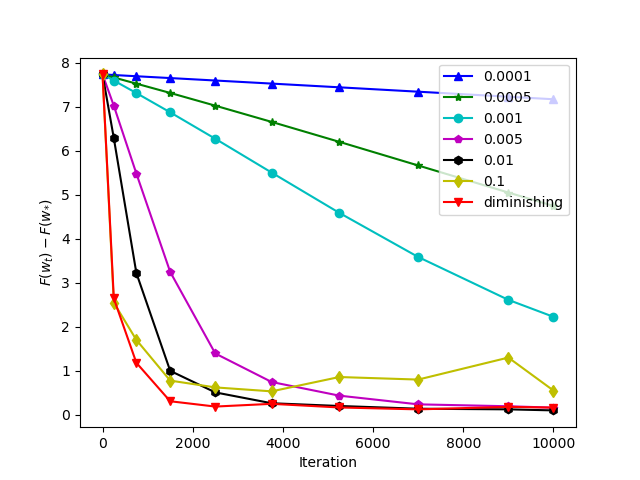}\label{fig:async_fl_stepsize_method_decay_vs_constant_w8a}}
  \caption{Asynchronous SGD (linearly increase sampling) with constant and diminishing step sizes.}
  \label{fig:async_fl_stepsize_method_decay_vs_constant}
\end{figure}

The detailed setup of this experiment is described in Table~\ref{tbl:tbl_async_fl_parameter}. In terms of diminishing step size, the experiment will start with the initial step size $\eta_{0}=0.1$. The experiment uses a linear increasing sample size sequence from round to round.

\textbf{Strongly convex case:}
Figure~\ref{fig:async_fl_strongly_convex_phishing_convergence} and Figure~\ref{fig:async_fl_strongly_convex_ijcnn1_convergence}  show that our proposed asynchronous SGD with diminishing step sizes and increasing sample size sequence achieves the  same or better accuracy when compared to asynchronous SGD with constant step sizes and constant sample sizes.
The figures depict strongly convex problems with diminishing step size scheme $\eta_t=\frac{\eta_0}{1 + \beta \cdot t}$ for an initial step size $\eta_{0}=0.1$ with a linearly increasing sample size sequence $s_i=a \cdot i^c + b$, where $c=1$ and $a, b \geq 0$ 
; diminishing$_1$ uses a more fine tuned  $\eta_t$ locally at the clients and diminishing$_2$ uses the transformation to round step sizes $\bar{\eta}_i$.
The number of communication rounds, see Figure~$\ref{fig:async_fl_phishing_convergence}$ as example, for constant step and sample sizes is $20$ rounds, while  the diminishing step size with increasing sample size setting only needs  $9$ communication rounds. 

\textbf{Plain convex case:} We repeat the above experiments for plain convex problems. Figure~$\ref{fig:async_fl_plain_convex_phishing_convergence}$ and Figure~$\ref{fig:async_fl_plain_convex_ijcnn1_convergence}$  illustrate the same results for the diminishing step size sequence $\eta_{t}=\frac{\eta_0}{1 + \beta \cdot \sqrt{t}}, \eta_0=0.1$ with increase sample size sequence  $s_i=a \cdot i^c + b$, where $c=1$ and $a, b \geq 0$.

\textbf{Non-convex case:} We run the experiment with the MNIST data set using the LeNet-$5$ model. We choose a diminishing step size sequence by 
decreasing the step size by $\eta_{t}=\frac{\eta_0}{1 + \beta \cdot \sqrt{t}}, \eta_0=0.1$ and use 
sample size sequence $s_i=a \cdot i^c + b$, where $c=1$ and $a, b \geq 0$. The detailed result of this experiment is illustrated in Figure~$\ref{fig:async_fl_mnist_neural_net}$.

In addition, for the strongly convex problems we extend our experiments to constant step size sequences, while still linearly increase the sample sizes $s_i=a \cdot i^c + b$, where $c=1$ and $a, b \geq 0$, from round to round.
The difference between using a  constant step size sequence plus increasing sample size sequence and diminishing step size sequence (starting at $\eta_0=0.01$) plus increasing sample size sequence can be found in Figure~\ref{fig:async_fl_stepsize_method_decay_vs_constant}. Overall, our asynchronous SGD with diminishing step sizes
gains good accuracy, which can only be achieved by fine tuning constant step sizes to $\eta=0.01$ and $\eta=0.005$ for the two data sets respectively. 

In conclusion, our proposed asynchronous SGD with diminishing step size sequences and increasing sample size sequences  works effectively for strongly convex, plain convex and non-convex problems because it can achieve the best accuracy when compared to other constant step size sequences while requiring fewer communication rounds.

\subsubsection{Asynchronous SGD with biased and unbiased data sets}
\label{subsec:dataset_FL_simu}

To study the behaviour of our proposed framework towards  biased and unbiased data sets, we run a simple experiment with in total $10,000$ iterations for $2$ compute nodes, where each compute node has its own data set.  The goal of this experiment is to find out whether our proposed asynchronous SGD can work well with biased data sets. Specifically, the first compute node will run for a data set which contains only digit 0 while the second client runs for a data set with only digit 1. Moreover, we choose the initial step size $\eta_{0}=0.01$ and a linearly increasing sample size sequence $s_i=a \cdot i^c + b$, where $c=1$ and $a, b \geq 0$ for strongly convex and plain convex problems. For simplicity, we choose  diminishing round step size sequence corresponding to $\frac{\eta_0}{1 + \beta \cdot t}$ for the strongly convex problem and $\frac{\eta_0}{1 + \beta \cdot \sqrt{t}}$ for the plain convex problem.

\begin{figure}[ht!]
  \centering
  \subfloat[Strongly convex.]{\includegraphics[width=0.5\textwidth]{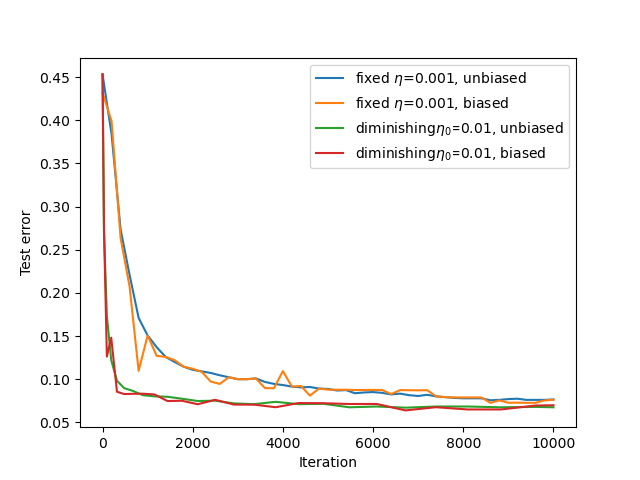}\label{fig:asyn_fl_biased_strongly_convex_mnist}}
  \hfill
  \subfloat[Plain convex.]{\includegraphics[width=0.5\textwidth]{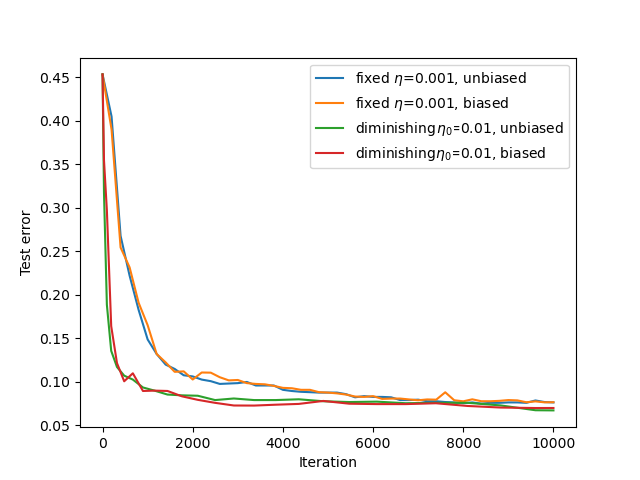}\label{fig:asyn_fl_biased_plain_convex_mnist}}
  \caption{Asynchronous SGD with biased and unbiased dataset (MNIST subsets)}
  \label{fig:asyn_fl_biased_mnist}
\end{figure}

\begin{figure}[H]
\begin{center}
\includegraphics[width=0.5\textwidth]{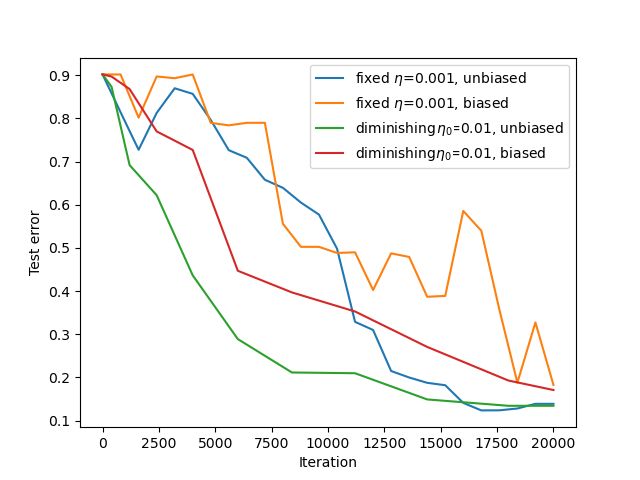}
\end{center}
\caption{Asynchronous SGD with biased and unbiased dataset (MNIST dataset)}
\label{fig:async_fl_biased_non_convex_mnist}
\end{figure}

As can be seen from Figure~$\ref{fig:asyn_fl_biased_mnist}$, generally, there is no significant difference when compute nodes run for biased or unbiased local data sets. This means that our proposed SGD framework can tolerate the issue of biased data sets, which is common in reality. Turning to the non-convex problem, we extend this experiment to the MNIST dataset, where the local data set of each compute node is separately biased, i.e, each of (the 5) compute nodes just has a separate subset of 2 classes of MNIST digits (hence, covering all $5\cdot 2$ digits). The experiment uses the initial step size $\eta_{0}=0.01$ with diminishing round step size sequence corresponding to $\frac{\eta_0}{1 + \beta \cdot \sqrt{t}}$.
Figure~$\ref{fig:async_fl_biased_non_convex_mnist}$ shows that while the accuracy might fluctuate during the training process, our asynchronous SGD still achieves good accuracy in general.

In conclusion, our asynchronous SGD framework  works well under biased data sets, i.e, this framework can tolerate the effect of biased data sets, which is quite common in reality.

\subsubsection{Asynchronous SGD with different number of compute nodes}
\label{subsec:num_worker}

We want to understand how varying the number  $n$ of compute nodes while fixing other parameters, such as the total number   of iterations $K=20,000$  and diminishing step size sequence, affects the accuracy. 
The goal of this experiment is to show that the number of clients $n$ can not be arbitrary large due to the restriction from the delay function $\tau$. 


To make the analysis simple, we consider 
asynchronous SGD 
with $d=1$ (i.e., each compute node is allowed to run faster than the central server for at most $1$ communication round) and unbiased data sets. Moreover, we choose a linearly increasing sample size sequence $s_i=a \cdot i^c + b$, where $c=1$ and $a=50, b=0$ for strongly convex and plain convex problems. The experiment uses an initial step size $\eta_{0}=0.01$ with diminishing round step size sequence corresponding to $\frac{\eta_0}{1 + \beta \cdot t}$ and $\frac{\eta_0}{1 + \beta \cdot \sqrt{t}}$ for strongly convex and plain convex problems respectively.

\begin{table}[ht!]
\caption{Test accuracy of asynchronous SGD with different number of compute nodes, strongly convex (phishing dataset)}
\label{tbl_async_fl_strongly_convex_worker}
\vskip 0.1in

\begin{center}
\begin{small}

\begin{tabular}{|c|c|c|}
\hline
\# of compute nodes & Accuracy (\%) & Duration (in second) \\ 
\hline
1         &   0.9355            &     338         
\\ 
\hline
2         &     0.9354          &     169         
\\ 
\hline
5         &     0.9297          &      57        
\\ 
\hline
10        &      0.9202         &      24        
\\ 
\hline
15        &      0.9134         &       17       
\\ 
\hline
20        &    0.9069           &      16       
\\ 
\hline
30        &    0.9005           &      16      
\\
\hline
\end{tabular}

\end{small}
\end{center}

\end{table}

\begin{table}[ht!]
\caption{Test accuracy of asynchronous SGD with different number of compute nodes, plain convex (phishing dataset)}
\label{tbl_async_fl_plain_convex_worker}
\vskip 0.1in

\begin{center}
\begin{small}

\begin{tabular}{|c|c|c|}
\hline
\# of compute nodes & Accuracy (\%) & Duration (in second) \\ 
\hline
1         &   0.9341         &   324     
\\ 
\hline
2         &   0.9303        &    164     
\\ 
\hline
5         &  0.9258       &     53   
\\ 
\hline
10        &  0.9247      &     26 
\\ 
\hline
15        &  0.9215       &     16     
\\ 
\hline
20        &  0.9135      &  15  
\\ 
\hline
30        &  0.9047         &     15
\\
\hline
\end{tabular}

\end{small}
\end{center}

\end{table}

As can be seen from Table~\ref{tbl_async_fl_strongly_convex_worker} and Table~\ref{tbl_async_fl_plain_convex_worker}, when we increase the number of compute nodes $n$, the training duration decreases gradually. Specifically, when $n=1$ (SGD with single machine) we achieve the best accuracy, compared to other settings. When $n=2$ or $n=5$, we get the same accuracy, compared to a single SGD setting while reducing the training time significantly. However, if we continue to increase  $n$ to a large number, for example $n=30$, then the accuracy has the trend to decrease and training duration starts to reach a lower limit. 

We first note that the lower accuracy is an artifact of our simulation: We split the training data set of size $M$ among each of the compute nodes (according to some random process). This means that each compute node uses its own $M/n$-sized local data set. The larger $n$, the smaller the local data sets, and as a result the local data sets are less representative of distribution ${\cal D}$ (the uniform distribution over the original $M$-sized training data set). In other words, the local distributions ${\cal D}_c$ start looking less and less like one another. This implies a shift from unbiased local data sets to biased local data sets for increasing $n$. This leads to a slight degradation in accuracy (see previous Section \ref{subsec:dataset_FL_simu}). If local data sets remain very large sample sets of the original training data, then we will keep on seeing the accuracy corresponding to unbiased local data sets (and this is what one would expect in practice).

Next we note that we are more likely to have a slow compute node among $n$ nodes if $n$ is large. The slowest compute node out of $n$ nodes is expected to be slower for increasing $n$. This means that other compute nodes will need to start waiting for this slowest compute node (see the while $\tau(t_{glob})\leq t_{delay}$ loop in \Call{MainComputeNode}{} which waits for the server to transmit a broadcast message once the slowest compute node has finished its round and communicated its update to the server). So, a reduction in execution time due to parallelism among a larger number $n$ of compute nodes will have less of an effect. For increasing $n$, the execution time (duration) will reach a lower limit. 

For larger $n$, the central server will process/aggregate a larger number of local updates -- nevertheless, since sample sizes increase from round to round, this should not become a bottleneck in later rounds (implying that compute nodes will not need to wait for the server finishing its computations/aggregations). 

In our simulations we see that a lower limit for the duration is reached for 15 compute nodes. We also see the same result for the experiment with  non-convex problem, which can be seen in Table~\ref{tbl_async_fl_non_convex_worker}. This experiment uses the initial step size $\eta_{0}=0.01$ with diminishing round step size sequence corresponding to $\frac{\eta_0}{1 + \beta \cdot \sqrt{t}}$.


As a final remark, if local data happens to be stored at many nodes in for example a data center, then our algorithm still scales to that setting: The compute nodes are used to bring computation to the data. They also compute in parallel but this will only reduce the full execution time/duration to a lower limit (after which parallelism will not further benefit a shorter full execution time). The increasing sample size sequence will still reduce communication with respect to  a constant sample size sequence or fixed sized mini-batch SGD.




\begin{table}[H]
\caption{Test accuracy of asynchronous SGD with different number of compute nodes, non-convex (MNIST dataset)}
\label{tbl_async_fl_non_convex_worker}
\vskip 0.1in

\begin{center}
\begin{small}

\begin{tabular}{|c|c|c|}
\hline
\# of compute nodes & Accuracy (\%) & Duration (in second) \\ 
\hline
1         &   0.9838        &     513        
\\ 
\hline
2         &   0.9815        &    403      
\\ 
\hline
5         &  0.9797      &     117  
\\ 
\hline
10        &   0.9177       &     77
\\ 
\hline
15        &  0.8809       &     75     
\\ 
\hline
\end{tabular}

\end{small}
\end{center}

\end{table}


\subsubsection{Asynchronous SGD with different number of iterations}
\label{subsec:iteration_T}

We now want to understand how varying the total number   of iterations $K=20,000$ while fixing other parameters, such as the number  $n$ of compute nodes  and diminishing step size sequence, affects the accuracy.
The goal of this experiment is to show that from some point onward it does not help to increase the number of iterations. This is because the test accuracy is measured with respect to a certain (test) data set of samples from distribution ${\cal D}$ for which the fraction of correct output labels is computed. Such a fixed test data set introduces an approximation error with respect to the training data set; the training accuracy which is minimized by minimizing the objective function is different from the test accuracy. 
Therefore, it does not help to attempt to converge closer to the global minimum than the size of the approximation error. Hence, going beyond a certain number of iterations will not reduce the estimated objective function (by using the test data set) any further.


To make the analysis simple, we consider 
asynchronous SGD 
with $d=1$ (i.e., each compute node is allowed to run faster than the central server for at most $1$ communication round) and unbiased data sets. Moreover, we choose a linearly increasing sample size sequence $s_i=a \cdot i^c + b$, where $c=1$ and $a=50, b=0$ for strongly convex and plain convex problems. The experiment uses an initial step size $\eta_{0}=0.01$ with diminishing round step size sequence corresponding to $\frac{\eta_0}{1 + \beta \cdot t}$ and $\frac{\eta_0}{1 + \beta \cdot \sqrt{t}}$ for strongly convex and plain convex problems respectively.
For simplicity, our simulation is based on $5$ compute nodes (together with the central aggregation server).

Our observation from Table~\ref{tbl_async_fl_num_iteration_strongly_convex},\ref{tbl_async_fl_num_iteration_plain_convex},\ref{tbl_async_fl_num_iteration_non_convex} is when the number of iterations is $50,000$, we gain the highest accuracy. In addition, if we continue to increase the number of iterations, the accuracy keeps nearly unchanged, i.e, a larger number of iterations does not improve accuracy any further.

\begin{table}[!ht]

\caption{Test accuracy of asynchronous SGD with different number of iterations, strongly convex (phishing dataset)}
\label{tbl_async_fl_num_iteration_strongly_convex}
\vskip 0.1in

\begin{center}
\begin{small}

\begin{tabular}{|c|c|c|}
\hline
\# of iterations & Accuracy (\%)  \\ 
\hline
1,000         &   0.9062          \\ 
\hline
2,000         &  0.9139              \\ 
\hline
5,000         &   0.9211            \\ 
\hline
10,000        &   0.9231              \\ 
\hline
20,000        &   0.9257            \\ 
\hline
50,000       &     0.9323           \\ 
\hline
100,000       &  0.9301              \\ 
\hline
\end{tabular}

\end{small}
\end{center}

\end{table}

\begin{table}[!ht]
\caption{Test accuracy of asynchronous SGD with different number of iterations, plain convex (phishing dataset)}
\label{tbl_async_fl_num_iteration_plain_convex}
\vskip 0.1in

\begin{center}
\begin{small}

\begin{tabular}{|c|c|c|}
\hline
\# of iterations & Accuracy (\%)  \\ 
\hline
1,000         &    0.9003              \\ 
\hline
2,000         &    0.9108            \\ 
\hline
5,000         &    0.9149           \\ 
\hline
10,000        &    0.9241             \\ 
\hline
20,000        &    0.9264             \\ 
\hline
50,000       &     0.9361           \\ 
\hline
100,000       &    0.9343            \\ 
\hline
\end{tabular}

\end{small}
\end{center}

\end{table}

\begin{table}[H]

\caption{Test accuracy of asynchronous SGD with different number of iterations, non-convex (MNIST dataset)}
\label{tbl_async_fl_num_iteration_non_convex}
\vskip 0.1in

\begin{center}
\begin{small}

\begin{tabular}{|c|c|c|}
\hline
\# of iterations & Accuracy (\%)        \\ 
\hline
1,000         &       0.4064        \\ 
\hline
2,000         &       0.7613        \\ 
\hline
5,000         &       0.9378        \\ 
\hline
10,000        &       0.9642        \\ 
\hline
20,000        &       0.9798        \\ 
\hline
50,000       &        0.9868        \\ 
\hline
100,000       &       0.9892        \\
\hline
\end{tabular}

\end{small}
\end{center}

\end{table}





\end{document}